\pdfoutput=1
\documentclass[11pt]{article}

\usepackage[preprint]{acl}

\usepackage{times}
\usepackage{latexsym}
\usepackage{tcolorbox}
\usepackage{svg}
\usepackage{multicol}
\usepackage[bottom]{footmisc}
\usepackage{stfloats}
\usepackage[shortlabels]{enumitem}

\usepackage{subcaption}
\usepackage{caption}  

\usepackage[T1]{fontenc}
\usepackage[utf8]{inputenc}

\usepackage{microtype}

\usepackage{inconsolata}
\usepackage{amsthm,amsmath,amssymb,amsfonts,amscd}
\DeclareMathOperator*{\argmin}{arg\,min} 
\DeclareMathOperator*{\argmax}{arg\,max} 
\DeclareMathOperator*{\argsmax}{args\,max} 

\title{Mass-Editing Memory with Attention in Transformers: A cross-lingual exploration of knowledge}

\author{
    Daniel Tamayo\textsuperscript{\rm 1} \quad 
    Aitor Gonzalez-Agirre\textsuperscript{\rm 1} \quad 
    Javier Hernando\textsuperscript{\rm 1,2} \quad 
    Marta Villegas\textsuperscript{\rm 1} \\
    \textsuperscript{\rm 1}Barcelona Supercomputing Center \\
    \textsuperscript{\rm 2}Universitat Politècnica de Catalunya \\
    \texttt{\{daniel.tamayo,aitor.gonzalez,javier.hernando,marta.villegas\}@bsc.es}
}

\begin{document}
\maketitle
\begin{abstract}

% Large language models excel at providing informed responses, drawing on internal knowledge. Current research examines how this knowledge resides within the model, focusing on methods like feed-forward weights and new information introduction. However, the role of attention in factual response generation remains nascent. This study investigates the cross-lingual capabilities of existing knowledge incorporation methods and furthers our understanding of attention's impact on factual accuracy. Drawing from these insights, we introduce MEMAT (Mass-Editing Memory with Attention in Transformers), a method that achieves significant improvements across all evaluation metrics through targeted parameter adjustments. MEMAT opens new avenues for enhancing factual response generation capabilities in large language models.

% Recent research has explored methods for updating and modifying information in large language models, often focusing on specific multi-layer perceptron blocks. This study expands on this work by examining the effectiveness of existing knowledge integration methods across languages and delving into the role of attention mechanisms in this process. Drawing from these insights, we propose MEMAT (Mass-Editing Memory with Attention in Transformers), a method that achieves significant improvements in all metrics while requiring minimal parameter modifications. MEMAT delivers a remarkable 10\% increase in magnitude metrics, benefits languages not included in the training data and demonstrates also a strong degree of portability.

Recent research has explored methods for updating and modifying factual knowledge in large language models, often focusing on specific multi-layer perceptron blocks. This study expands on this work by examining the effectiveness of existing knowledge editing methods across languages and delving into the role of attention mechanisms in this process. Drawing from the insights gained, we propose Mass-Editing Memory with Attention in Transformers (MEMAT), a method that achieves significant improvements in all metrics while requiring minimal parameter modifications. MEMAT delivers a remarkable 10\% increase in magnitude metrics, benefits languages not included in the training data and also demonstrates a high degree of portability. Our code and data are at \href{https://github.com/dtamayo-nlp/MEMAT}{https://github.com/dtamayo-nlp/MEMAT}.

\end{abstract}

\section{Introduction}
w
Large Language Models (LLMs) based on transformers \cite{AttIsAllYouNeed} are designed to predict the probability of tokens occurring in a sentence rather than comprehending the true semantics that underlie it. As a result, they are susceptible to generating content that lacks a solid grounding in reality and accuracy. Even when two prompts relate to the same factual association $\left<s,r, \cdot\right>=\left< \text{\textit{Google}}, \text{\textit{CEO}}, \cdot \right>$; ``\textit{The CEO of Google is}” and ``\textit{Google's CEO is}”, the model lacks an internal constraint that compels it to generate identical answers.

Different investigations have already highlighted the limitations of the models’ genuine understanding by analyzing its dependency on the dataset patterns \cite{annotationArt,adversarial_examples}. Furthermore, even when these models seem to \textit{know} the correct answer to a given prompt, they exhibit vulnerability when provided harmful context \cite{overthinking}. 

In an initial pursuit of exploring the model's true understanding when using knowledge editors, we analyze Mass-Editing Memory in Transformers (MEMIT) \cite{MEMIT}, a knowledge-editing method asserting its capability to insert up to 10,000 factual associations without heavily inducing catastrophic forgetting. Aligned with previous research \cite{CrossLingual}, our first investigation involves a cross-lingual examination of the limitations associated with MEMIT.

Although the cross-lingual consistency is dependent on the similarity between languages \cite{clc}, our study specifically delves into examining the polyglot capabilities between English and Catalan. In this segment, we construct a translation pipeline to mitigate differences between these languages and proceed to investigate the impact of subject tokenization on knowledge incorporation. 
% It should be noted that due to the significant similarity between entities in English and Catalan, the findings presented may not be applicable to languages exhibiting lower levels of resemblance.

Motivated by the potential of language-independent knowledge neurons \cite{JourneyCenterKnowledge}, and the relevance of the attention mechanism in the factual associations domain \cite{DissectingRecall}, we further our study by exploring a particular part of the attention mechanism: the attention heads. Attention heads have proven to be useful in enhancing the model's reliability under Inference-Time Intervention (ITI) \cite{ITI}. The foundational hypothesis behind ITI suggests that attention heads serve as key sources of information for evaluating the truthfulness of models when presented with sentences. Through our experiments, we not only validate the extension of this claim to the domain of factual associations but also observe promising outcomes from a cross-lingual lens. Building on these insights, we propose MEMAT, a method that introduces a novel approach to guide the model towards a better understanding of the edited factual associations.

The proposed method demonstrates improvement across all evaluation metrics from both cross-lingual and monolingual perspectives, showcasing differences exceeding 10\% in some cases. Furthermore, additional experiments suggest that the modifications introduced by our algorithm enhance the model's understanding of existing knowledge rather than reintroducing it, rendering this approach portable and computationally efficient.

\section{Related Work}

\textbf{Retrieval Methods.} Rather than directly relying in LLMs for specific queries, open-domain question answering systems has historically been driven by the development of algorithms aligning queries with external database sources \cite{BM25}. Recent advancements in aligning retrieval-based methods with LLMs have demonstrated promise in this domain \cite{dense_passage,GAR,RETRO}, with retrieval augmented generation \cite{RAG} showing capabilities in both multimodal \cite{Chen_multimodal,Chen_multimodal2,RAGmultimodal} and multilingual \cite{retrievalMulti} contexts. However, while the use of external sources avoids the need for fine-tuning, challenges still persist in precisely identifying the relevant context for a given query \cite{RAG_surv}.

\textbf{Truthfulness.} Efforts to enhance the reliability of LLMs without depending on external sources have been a focal point of recent research. Aligning LLMs with human feedback has been explored through Reinforcement Learning from Human Feedback \cite{RLHF2,RLHF} and Direct Preference Optimization \cite{DPO}, offering valuable insights for veracity alignment \cite{GRATH}. Additionally, approaches contrasting hidden representations of these models have also yielded significant results \cite{Contrastive_L,DOLA} in this direction.

\textbf{Factual Knowledge Editors.}
This research builds upon MEMIT, a method adept at efficiently introducing knowledge by modifying the internal weights of decoder-only architectures, surpassing the effectiveness of earlier meta-learning techniques like MEND \cite{MEND} and constrained fine-tuning \cite{FT_normal}. Nevertheless, less intrusive alternatives, which selectively modify specific hidden states of the model during inference according to the provided prompt, have also demonstrated remarkable efficacy in knowledge editing. Notable examples include REMEDI \cite{REMEDI}, GRACE \cite{Aging}, and SERAC \cite{serac}. 

\textbf{Multilingual Domain.}
The emergence of knowledge editors and multilingual models raises questions about whether the information is being inserted from a cross-lingual perspective. Current findings suggest that these methods are not entirely language-independent \cite{PolyglotOrNot,CrossLingual}, with approaches based on prompting and retrieval yielding stronger results \cite{IKE,retrievalMulti}. 

% Nevertheless, concurrent works that point to the existence of language-independent knowledge neurons \cite{JourneyCenterKnowledge} and masking-based approaches \cite{LanguageAnisotropic} hint at potential solutions to the cross-lingual knowledge editing problem.

% \cite{ripple_effects}
% \cite{DOLA} (Contrastive)

\section{Preliminaries}

\subsection{Background}
Since in our experimental setup English and Catalan were chosen as the languages for conducting experiments, we opted for the utilization of Ǎguila-7B, a decoder-only model consisting of 6.85 billion parameters based on Falcon-7B \cite{falcon}. The internal process performed by this architecture to process text is similar to other decoder-only architectures. It first convert an input to a sequence of $N$ tokens $t_1,t_2,...,t_N$ by using Byte-level Byte-Pair Encoding \cite{BBPE}. Then, it process each token by assigning a vector $x_i^{0}$ using an embedding matrix $E\in \mathbb{R}^{|\mathcal{V}|}\times d$, where $\mathcal{V}$ denotes the set of vocabulary tokens and $d$ denotes the size of each vector. Following this, the input embeddings undergo a series of L transformer layers, each comprising a Multi-Query Self-Attention (MQSA) sublayer \cite{MultiQueryAttention} and a parallel Multi-Layer Perceptron (MLP) sublayer. 

Following the notation proposed in \citet{Circuits,DissectingRecall}, we avoid representing bias terms, layer normalization \cite{layerNorm}, and Rotary Position Embeddings \cite{Rope} for simplicity and denote the transformation as:
\begin{equation}
    x_i^{\ell} = x_i^{\ell-1} + a_i^{\ell} + m_i^{\ell},
\end{equation}

where $a_i^{\ell}$ and $m_i^{\ell}$ are the outputs from the $\ell$-th MQSA and MLP sublayers. In the attention term, for each layer, we assign different projection matrices $W_Q^{\ell,h}, W_K^{\ell}, W_V^{\ell}\in \mathbb{R}^{d\times \frac{d}{H}}$ and $W_O^{\ell,h} \in \mathbb{R}^{\frac{d}{H}\times d}$ for $h\in [1,H], \ell \in [1,L]$. Then, given the hidden states of the sentence at layer $\ell$ denoted as $X^{\ell} \in \mathbb{R}^{N\times d}$, we define:
\begin{align}
    A^{\ell,h} & = \mathcal{S} \left(\frac{(X^{\ell-1}W_Q^{\ell,h})(X^{\ell-1}W_K^{\ell})^T}{\sqrt{d/H}}+M^{\ell,h}\right) \label{eq:attn_m} \\
    a^{\ell} & = \sum_{h=1}^{H} A^{\ell,h}(X^{\ell-1}W_V^{\ell})W_O^{\ell,h} \label{eq:attn_term},
\end{align}

where $\mathcal{S}$ is a row-wise softmax normalization, and $M^{\ell,h}$ is a mask for $A^{\ell,h}$ that only uses the attention mechanism to modify the token $t_r$ using the previous tokens $t_{\leq r}$  ($M^{\ell,h}_{rc}=-\infty$ $\forall c > r$ and zero otherwise).

In the MLP term, we use the matrices $W_{in}\in \mathbb{R}^{d\times d_{ff}}$, $W_{out}\in \mathbb{R}^{d_{ff}\times d}$ and an activation function $\gamma$ to define:
\begin{gather}
    \begin{split}
        k_i^{\ell} & =\gamma(x_i^{\ell-1}W_{in}) \\
        m_i^{\ell} & = k_i^{\ell} W_{out}.
        \label{eq:ff}
    \end{split}
\end{gather}

\subsection{Proposed Framework}

While the architecture of large language models is extensively documented, grasping the precise mechanisms that empower them to extract factual information is still matter of research. Notably, studies have revealed the impact of adjusting MLP layers in the generation of factual associations \cite{geva-key-value,DAIknowledge,JourneyCenterKnowledge}. This comprehension has paved the way for the development of frameworks such as ROME \cite{ROME}, PMET \cite{PMET}, and MEMIT. In the case of PMET and MEMIT, a subset of MLP layers are changed by the introduction of a correction matrix ($\widehat{W}_{out,\ell} = W_{out,\ell}+ \Delta_{\ell}$) such that:
\begin{gather}
\begin{split}
     \widehat{W}_{out,\ell} &= \argmin_{\widetilde{W}_{out,\ell}} (\sum_{j=1}^n ||k_j^{\ell} \widetilde{W}_{out,\ell} -\widetilde{m}_j^{\ell}||^2+ \\
     & \sum_{j=n+1}^{n+u}||k_j^{\ell} \widetilde{W}_{out,\ell}-\widetilde{m}_j^{\ell}||^2).
\end{split}\label{eq:MLP_intro}
\end{gather}
where $n$ represents the number of factual associations already encoded in the pre-trained model, $u$ represents the number of new factual associations being introduced, each $k_i^{\ell}$ is taken from the final position of the subject entities of each factual triplet, and the representation of $\widetilde{m}_j^{\ell}$ is the one that should be capable of making the model predict the correct factual entity. Refer to \citet{MEMIT} for more details.

Despite the notable performance of these proposals, recent studies highlight the critical role of attention mechanisms in accurate response generation \cite{whyInContextLearning,attention_satisfies}. Specifically, its relevance in factual associations when the attribute extraction is performed \cite{DissectingRecall}. These findings have already prompted some intervention of attention layers for knowledge editing \cite{PMET,multihop}, a line of research that this study aims to further extend by tailoring attention with the ITI framework. 

The core principle of ITI involves a simple method for calculating a subset of head corrections ($\omega^{\ell, h}$) that, when integrated into the language model:
\begin{align}
    \tilde{a}^{\ell} & = \sum_{h=1}^{H} (A^{\ell,h}(X^{\ell-1}W_V^{\ell})+\omega^{\ell, h})W_O^{\ell,h},
\label{eq:MEMAT_e}\end{align}
result in a significant enhancement in the veracity of the model's responses.

Nevertheless, as knowledge is infused through MEMIT, the notion of truth becomes nuanced. The model can express new information in specific contexts, yet upon closer examination of its reasoning capabilities, a decline in performance is observed \cite{ripple_effects}. The distinctive approach introduced in MEMAT explores how incorporating modifications based on head corrections can optimize the method's understanding of the knowledge introduced while avoiding catastrophic forgetting. Refer to Section \ref{sec:MEMAT} for further details. 

% Truthfulness shoud

% The acknowledged hypothesis on attention's role focuses on attribute extraction phenomena \cite{DissectingRecall}. Upon scrutinizing experimental results in the new framework, we posit that not all heads might be pertinent to accurately perform the attribute extraction of factual associations and that due to its relation with truthfulness, their usefulness for a cross-lingual lens might be superior than multi-layer percetrons.

% \newpage
\section{Dataset and Evaluation}

The dataset employed in this document is a reduced version of the CounterFact dataset \cite{ROME, MEMIT}. For each sample, the relevant prompts utilized in our experiments are:
\begin{enumerate}
    \item \textit{Efficacy Prompts (EP)}. Two distinct objects associated with the same $\left< s,r,\cdot \right>$ pair, one corresponding to the true fact $o^c$ and the other representing a false fact $o^*$. 
    \item \textit{Paraphrase Prompts (PP)}. Two prompts that have the same meaning of the $\left< s,r,\cdot \right>$ pair, but are paraphrased and receive an addition of noise at the beginning. In evaluations, these prompts can also be referred to as indicators of the model's \textit{generalization} capability.
    \item \textit{Neighborhood Prompts (NP)}. Ten different prompts which contain different subjects ($s_j\neq s$) with the same relation $\left< s_j,r,\cdot \right>$ that would be true with the object $o^c$.  In evaluations, these prompts are referenced as indicators of the model's ability to \textit{specify} the insertion of knowledge.
\end{enumerate}

As discussed in \citet{PolyglotOrNot}, a purification of the original dataset is necessary to eliminate sentences with awkward phrasing, consistent inaccuracies, and errors. After performing the refinement proposed, the resulting English dataset consists of 11,550 factual associations. However, since our goal is also to investigate the cross-lingual capabilities of MEMIT and there is no available translated dataset meeting our criteria, we develop a translation pipeline. 

In contrast to the methodologies employing Wikidata and Google knowledge graphs for translation verification \cite{mLAMA}, and utilizing \texttt{gpt-3.5-turbo} and \texttt{gpt-4} for translations \cite{CrossLingual}, our implementation follows a distinct procedure. Our pipeline translates English sentences to Catalan using \href{https://huggingface.co/projecte-aina/mt-aina-en-ca}{\texttt{projecte-aina/mt-aina-en-ca}}, and maintains the dataset structure with \texttt{simalign} \cite{simalign}, an aligner method based on contextual embeddings. 

Given the significance of preserving sentence order, cases in which the targets $o^c$ or $o^*$ are translated before the end of the sentences are flagged as errors and discarded if necessary. Additionally, we address a challenge associated with gender differences between English and Catalan. For instance, if the English sentence ``\textit{The CTO of OpenAI is}” is translated as ``\textit{El CTO d'OpenAI és}”, it may introduce bias toward male responses even when the correct answer is $o^c=\text{\textit{Mira Murati}}$. To mitigate this, we create two Catalan samples for each English sample.

Using our pipeline and human supervision, we manage to obtain a reduced version of the CounterFact dataset in English and Catalan containing 11,229 samples. An example of the samples in Catalan can be observed in Appendix \ref{appendix:counterfact}.

% \subsection{Evaluation}
To evaluate the performance of the different knowledge editors used in our experiments, this study inherits the evaluation metrics from \citet{ROME, MEMIT}. For each of the three different prompts contained in the dataset (EP, PP, NP), success and magnitude metrics are defined:

\begin{itemize}
    \item Success Metrics:
\begin{gather}
    \begin{split}
    \text{ES} & := \mathbb{E} \left[\mathbb{P}[o^*|p]>\mathbb{P}[o^c|p] | p\in EP \right] \\
    \text{PS} & := \mathbb{E}\left[\mathbb{E}_{p\in PP} \left[\mathbb{P}[o^*|p]>\mathbb{P}[o^c|s]\right] \right]\\
    \text{NS} & := \mathbb{E}\left[\mathbb{E}_{p\in NP} \left[\mathbb{P}[o^c|p]>\mathbb{P}[o^*|s]\right] \right].
    \end{split}
\end{gather}
    \item Magnitude Metrics:
\begin{gather}
    \begin{split}
    \text{EM} & := \mathbb{E} \left[\mathbb{P}[o^*|p]-\mathbb{P}[o^c|p] | p\in EP \right] \\
    \text{PM} & := \mathbb{E}\left[\mathbb{E}_{p\in PP} \left[\mathbb{P}[o^*|p]-\mathbb{P}[o^c|s]\right] \right]\\
    \text{NM} & := \mathbb{E}\left[\mathbb{E}_{p\in NP} \left[\mathbb{P}[o^c|p]-\mathbb{P}[o^*|s]\right] \right].
    \end{split}
\end{gather}
\end{itemize}

% \begin{itemize}
%     \item Accuracy Metrics:
%     \begin{gather}
%     \begin{split}
%         \text{EA} &:= \mathbb{E}  \left[o^* = \argmax_{\omega} \mathbb{P}[\omega|p] | p\in EP\right] \\
%         \text{PA} &:= \mathbb{E}\left[\mathbb{E}_{p\in PP} \left[o^* = \argmax_{\omega} \mathbb{P}[\omega|p]\right] \right] \\
%         \text{NA} &:= \mathbb{E}\left[\mathbb{E}_{p\in NP} \left[o^c = \argmax_{\omega} \mathbb{P}[\omega|p]\right] \right].
%     \end{split}
%     \end{gather}
% \end{itemize}

In prior research \cite{ROME, MEMIT,PMET,CrossLingual}, the assessment of knowledge editors heavily relied on success metrics. Nevertheless, it is crucial to note that a favorable success metric coupled with a low-magnitude metric could suggest uncertainty in the model's confidence regarding the retrieved knowledge. In the following sections, we emphasize the significance of the magnitude metric in uncovering patterns that may not be readily apparent in traditional success metrics.

% \newpage
\section{Experiments}
Before the introduction of MEMAT, the main aspects that motivate the use of our method are explained in this section. Firstly, Section \ref{sec:cross_ling} outlines the scope of our analysis and examines the limitations of using only English and Catalan. Our findings suggest a correlation between positive cross-lingual outcomes in MEMIT and higher token similarity between subject tokens, indicating the dependency of our cross-lingual analysis on languages that share subject tokens. Subsequently, in Section \ref{sec:loc}, we evaluate the extent of cross-lingual information in the hidden representations of words by studying attention heads.

As Ǎguila-7B was not initially assessed using MEMIT, further details on the hyperparameter optimization of the methods studied can be found in Appendix \ref{appendix:optimiz}.

\subsection{Cross-Linguality}\label{sec:cross_ling}

\begin{figure}[htb!]
  \centering
  \begin{subfigure}[b]{0.5\textwidth}
    \caption{Success Metrics}
    \centering
    \includegraphics[width=7 cm, trim= 0cm 2.3cm 0cm 0cm, clip]{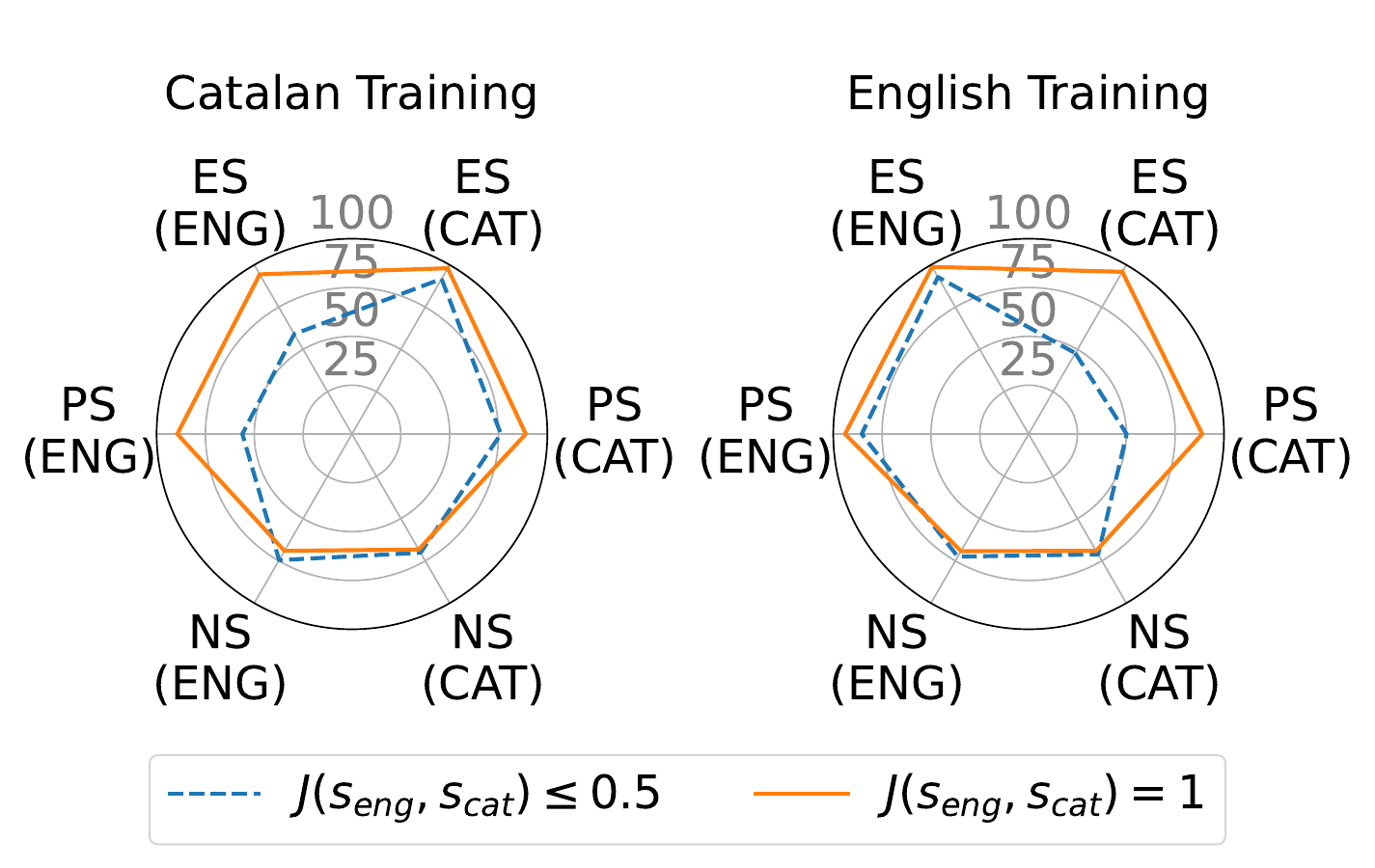}
  \end{subfigure}
  
  % \begin{subfigure}[b]{0.5\textwidth}
  %   \caption{Accuracy Metrics}
  %   \includegraphics[width=\textwidth, trim= 0cm 0cm 0cm 0cm, clip]{radar_charts/Acc.pdf}
  % \end{subfigure}

  \begin{subfigure}[b]{0.5\textwidth}
    \caption{Magnitude Metrics}
    \centering
    \includegraphics[width=7 cm]{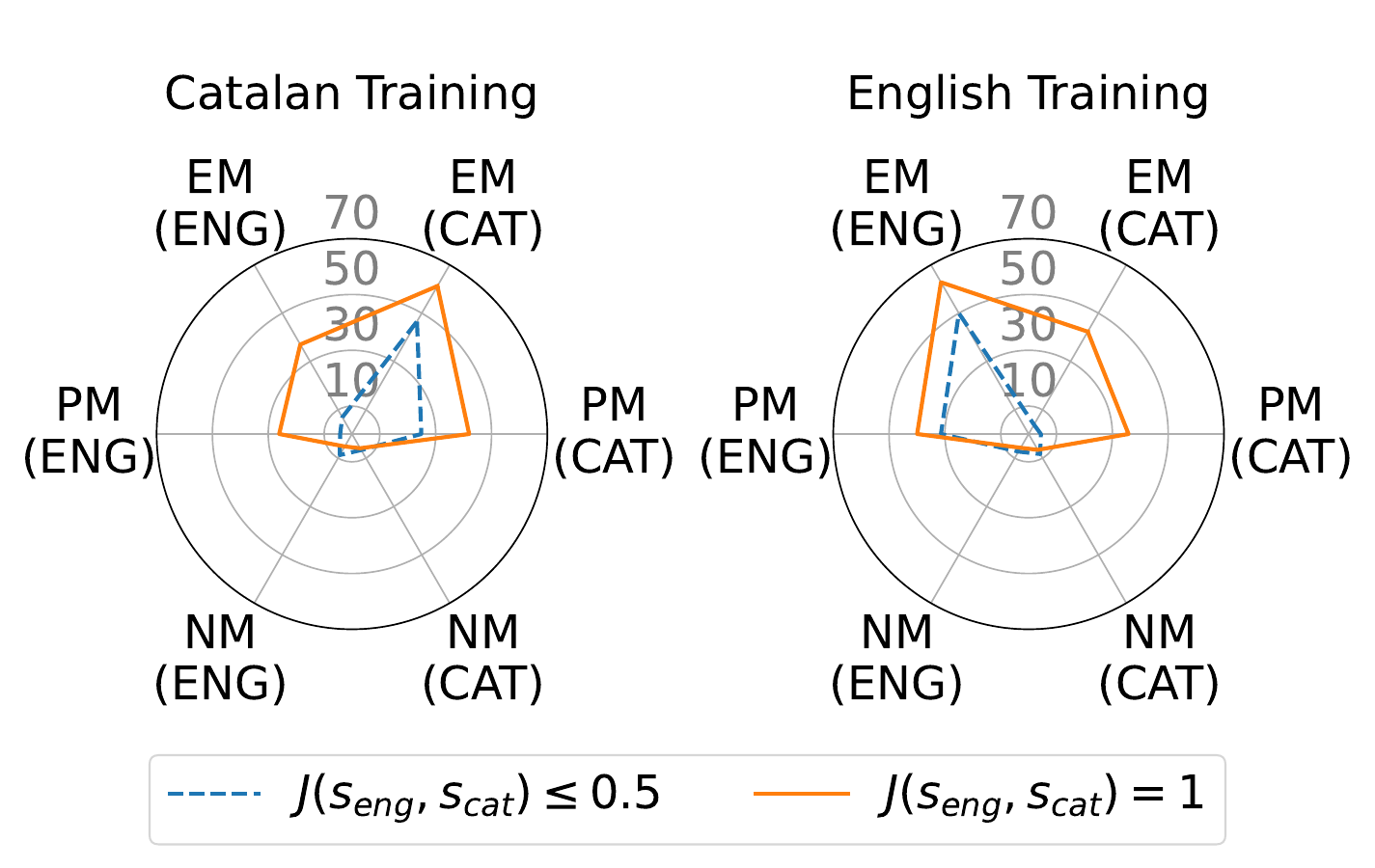}
  \end{subfigure}
  \caption{Results of Efficacy, Generalization and Specificity when applying MEMIT separately in two different languages and evaluating the effects of training in both. Each depicted line show a restriction in the tokenization of the subjects. }
  \label{fig:bilingual}
\end{figure}

Considering the substantial resemblance between the English and Catalan alphabets,  we investigate the impact of this similarity on the cross-lingual hypotheses asserted in this paper. Utilizing the Jaccard index, expressed as:
\begin{equation}
    J(x_{eng},x_{cat}) = \frac{|x_{eng}\cap x_{cat}|}{|x_{eng}\cup x_{cat}|},
\end{equation}\label{eq:jaccard}
\noindent
we assess the performance disparity when incorporating factual triplets with distinct subject tokenizations in English and Catalan (\(J(s_{eng}, s_{cat})\leq 0.5\)), as opposed to those without such differences (\(J(s_{eng}, s_{cat})= 1\)). It is pertinent to note that factual associations typically pertain to entities such as institution names, individuals, and series, which tend to maintain consistent tokenization across languages that share the same alphabet. More details of the exact similarity between both datasets can be found in Appendix \ref{appendix:sim}.

In Figure \ref{fig:bilingual}, the outcomes of cross-lingual operations are illustrated for the insertion of 1,000 samples using MEMIT. Two discernible trends emerge from the results:
\begin{itemize}
    \item Given the dependency of MEMIT in the subject representation, alterations to the subject result in a more pronounced decline in performance from a cross-lingual perspective. This phenomenon may explain the less cross-lingual outcomes observed in Chinese \cite{CrossLingual}.
    \item When analyzing cases with the same tokenization $(J(s_{eng}, s_{cat})= 1)$ in both languages, the decline in cross-lingual magnitude metrics is more noticeable than the decrease observed in success metrics. Instances with $(J(s_{eng}, s_{cat})\leq 0.5)$ experience a significant decrease in performance across both metrics.
\end{itemize}

\begin{figure}[b!]
  % \centering
  \begin{subfigure}[b]{0.5\textwidth}
    \caption{$L_1=$ Catalan, $L_2=$ Catalan}
    % \centering
    \includegraphics[width=7.7 cm,trim= 0cm 1.6cm 0cm 3cm, clip]{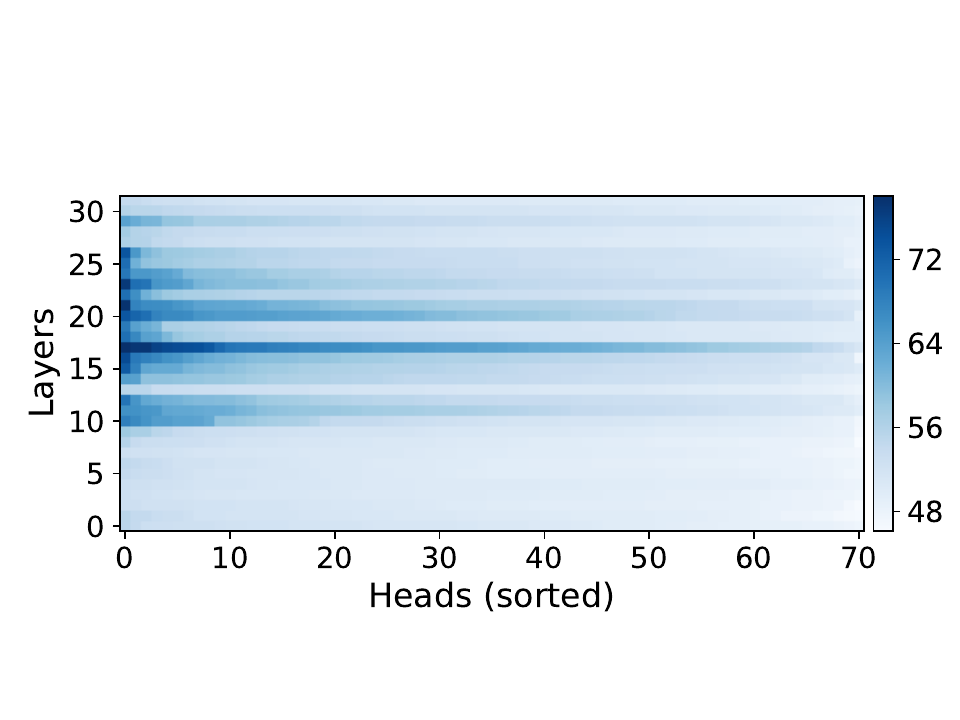}
  \end{subfigure}
  \begin{subfigure}[b]{0.5\textwidth}
    \caption{$L_1=$ Catalan, $L_2=$ English}
    % \centering
    \includegraphics[width=7.7 cm,trim= 0cm 1.6cm 0cm 3cm, clip]{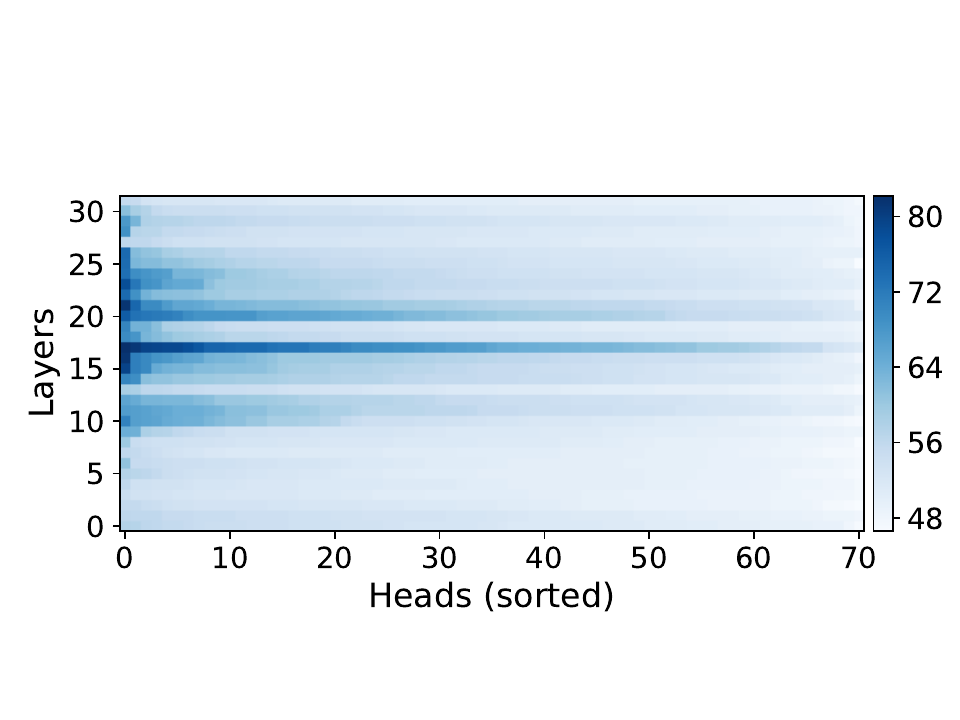}
  \end{subfigure}
  \caption{Accuracy on the validation set for all heads in all layers in Ǎguila-7B considering two combinations of $L_1$ and $L_2$. The performance peaks include 78.1\% and 82.2\%. The number of samples introduced using MEMIT is 1,000.}
  \label{fig:heads}
\end{figure}
\subsection{Locating Knowledge with Heads}\label{sec:loc}
Under the cross-lingual context outlined in the previous section, we analyze the extent to which the framework of ITI can be useful in the factual knowledge domain. Considering that we have the same set of factual associations in two languages, we denote a language pair as $(L_1, L_2)$ and design the location of knowledge using a specific part of the attention mechanism as follows:

\begin{enumerate}
    \item We train the model using MEMIT with triplets on language $L_1$.
    \item We identify all heads associated to the last token of $M$ the triplets $\left<s, r, o^c \right>_i$ and $\left<s, r, o^* \right>_i$ using $L_2$, assigning truthful labels $y=0$ and $y=1$ respectively\footnote{Note that the concept of \textit{truth} in this case is diffuse since the model has already been trained on $L_1$ and the new true target should be $o^*$.}. The constructed dataset has the structure: $\{(head^{\ell, h}_{-1,\cdot}, y)_i\}^{M}_{i=1}$, where each head of each layer can be denoted as:
    \begin{equation}
        head^{\ell, h} = A^{\ell,h}(X^{\ell-1}W_V^{\ell}).
    \end{equation}
    \item We train sigmoid classifiers for each head position (totaling $L \times H$ positions) on subset of triplets, denoted as the training set. The objective is to predict the assigned label by just using a single attention head representation. Subsequently, we utilize the remaining triplets as a validation set to assess the performance of this approach. More implementation details can be found in Appendix \ref{appendix:details}.
\end{enumerate}

If the attention heads could not provide information about whether the sentences are truthful or not, the expected performance should be around a 50\% chance of predicting the correct label. However, as highlighted in \citet{ITI}, empirical observations in a comparable context revealed that certain attention heads achieved 83.3\% performance on the validation set in discerning truthful sentences.
\begin{figure*}[b!]
    \centering
    \includegraphics[width=1\textwidth, trim= 0cm 7.6cm 16.5cm 1cm, clip]{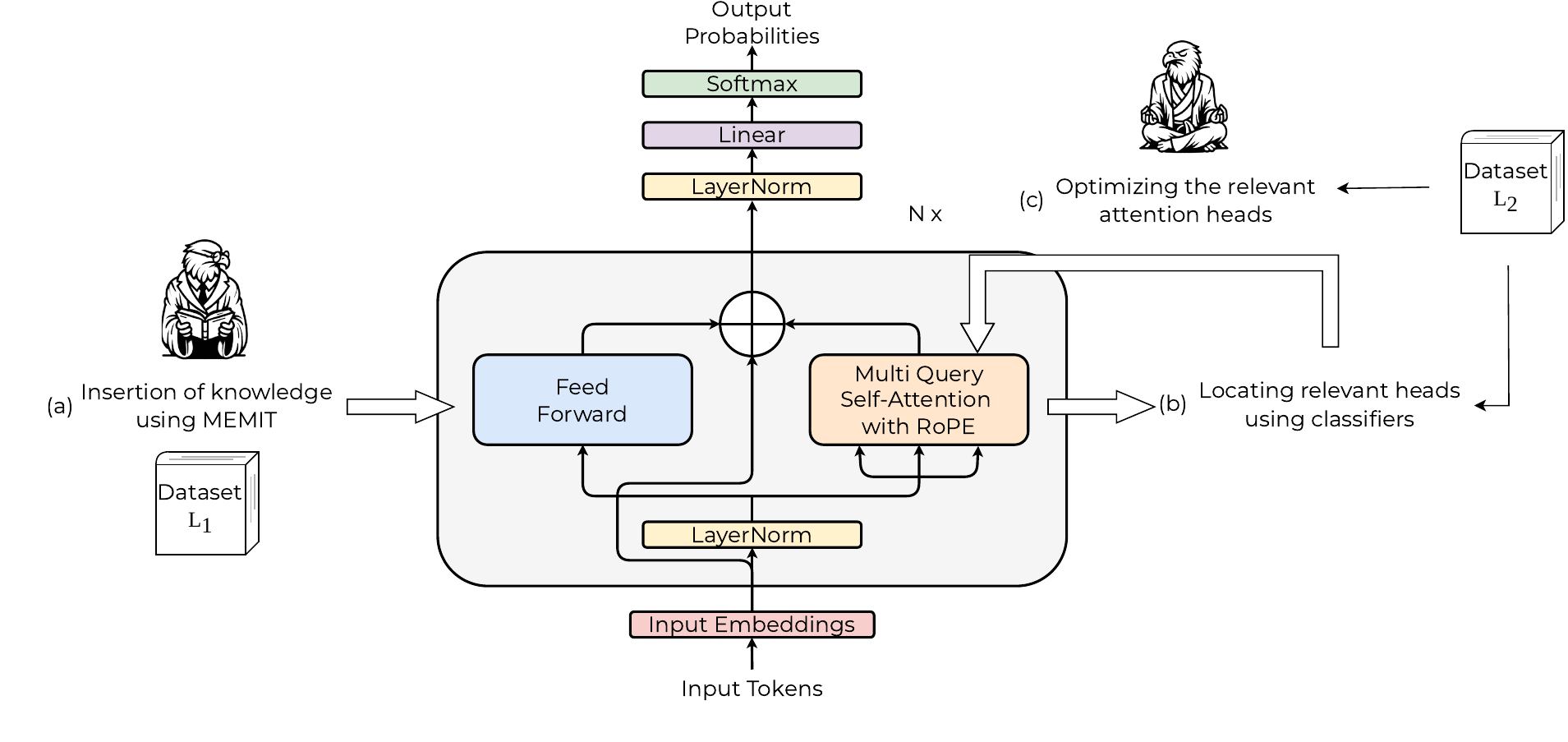}
    \caption{Illustration depicting the key steps of MEMAT in Ǎguila-7B. The dataset languages, denoted as $L_1$ and $L_2$, are not restricted to differ or remain equal, but in this diagram we consider both datasets to store the same triplets. The Eagle images were generated using GPT-4.}
    \label{fig:MEMAT}
\end{figure*}
In this section, our contribution involves not only expanding the application of this framework to MEMIT, but also demonstrating a high degree of language independence. Regardless of the choice of languages $L_1$ and $L_2$, some attention heads consistently achieve accurate classification performances near 80\%, as shown in Figure \ref{fig:heads}. 

Note that we are just exploring the cross-lingual implications from Catalan to English, but similar patterns can be observed in Appendix \ref{appendix:heads} for the converse relation.

\section{MEMAT Method}\label{sec:MEMAT}

In light of the proven attention heads' relevance, we reinforce the rationale behind equation \ref{eq:MEMAT_e} and present MEMAT as a method that expands upon MEMIT. The overall procedure is depicted in Figure \ref{fig:MEMAT}, with detailed descriptions for each point as follows:

\begin{enumerate}[(a)]
    \item Firstly, we modify the model with knowledge associated to a set of factual triplets using MEMIT in language $L_1$, which only edit some MLP layers.
    \item Then, using language $L_2$, we locate the heads that yield the top $K$ performances using the procedure explained in Section \ref{sec:loc}. Formally, let us consider that, for each classifier learned using the training set, we obtain the predictions $\phi_i^{\ell, h}=H( < head_{-1,\cdot}^{\ell, h}, \theta^{\ell, h} >)$, where $H$ denotes the Heaviside step function and the parameters $\theta^{\ell, h}$ have been trained in the training set. The top head positions can be denoted as those which belong to the set:
    \begin{gather}
    \begin{split}
        & \Psi^K  := \{(\ell,h) |\argsmax_{(\ell,h)}^{K} ( \{ \phi_i^{\ell, h}\wedge y_i\}_{i=1}^{M\times \beta })\},
    \end{split}
    \end{gather}
    where $\beta$ is the fraction of the validation set.
    \item Finally, under the language $L_2$, we introduce head corrections $\omega^{\ell,h}$ in each of the $K$ head positions, $\Psi^{K}$, and minimize the loss function:
    \begin{gather}
    \begin{split}
         & \mathcal{J}_i^{attn} = \frac{\lambda_{\omega }}{K} \sum_{(\ell,h)\in \Psi^K}\left( \frac{||\omega^{\ell, h}||}{||head_{-1,\cdot}^{\ell, h}||}\right)^2 \\
         & - \frac{1}{R}\sum_{j=1}^R \log \mathbb{P}_{\widetilde{G}}[o^*_i|z_j + p(s_i,r_i)]\\
         & + D_{KL}\left(\mathbb{P}_{\widetilde{G}}[x|p']||\mathbb{P}_G [x|p']\right),
    \end{split}\label{eq:loss_attn}
\end{gather}
\end{enumerate}
% \min_{\left[\substack{\omega^{\ell, h} \\ (l,h)\in \Psi^K}\right]}

\begin{itemize}[label={}]
  \item where $\widetilde{G}$ represents the modified decoder model obtained by inserting $\omega^{\ell, h}$ in the decoder $G$ that results from point (a), ($\widetilde{G} =G([head_{-1,\cdot}^{\ell, h}]+=[\omega^{\ell, h}])$). The term $D_{KL}$ is the KL divergence that minimize the effect of the modification in prompts $p'$ that contain the subject $s_i$, but contain the relationship "\textit{is a}". Finally, $R$ is the number of random prompts $z_j$ that are inserted at the beginning of the sentence to make the optimization more robust under different contexts. 
  
  Once the head corrections are optimized, we apply equation \ref{eq:MEMAT_e}.
\end{itemize}

\begin{table*}[ht!]
% \resizebox{\textwidth}{!}{
\begin{tabular}{ccccccc}
\hline
\begin{tabular}[c]{@{}c@{}}Method \\ (Training Language(s))\end{tabular} & \begin{tabular}[c]{@{}c@{}}English\\ ES\end{tabular} & \begin{tabular}[c]{@{}c@{}}Catalan\\ ES\end{tabular} & \begin{tabular}[c]{@{}c@{}}English\\ PS\end{tabular} & \begin{tabular}[c]{@{}c@{}}Catalan\\ PS\end{tabular} & \begin{tabular}[c]{@{}c@{}}English\\ NS\end{tabular} & \begin{tabular}[c]{@{}c@{}}Catalan\\ NS\end{tabular}   \\ \hline
Ǎguila-7B Baseline & 26.5 (1.2) & 25.5 (1.2) & 30.4 (1.2) & 31.6 (1.2) & \textbf{73.9 (0.9)} & 72.2 (0.8) \\ \hline
PMET (CAT) & 80.4 (2.5) & 95.6 (1.3) & 72.5 (2.5) &81.3 (2.3) & \textbf{73.9 (2.1)} &\textbf{74.3 (1.8)} \\ \hline
MEMIT (CAT)                                                           & 88.8 (0.7)                                           & 97.4 (0.4)                                           & 84.0 (0.8)                                           & 88.3 (0.7)                                           & 71.8 (0.8)                                           & 70.5 (0.7)\\ \hline
MEMAT-16 (CC)                                                         & 90.3 (1.1)                                          &\textbf{97.8 (0.5)}                                           & 86.2 (1.1)                                           & 89.6 (1.0)                                           & 72.7 (1.2)                                           & 71.8 (1.0) \\ \hline
MEMAT-16 (CE)                                                         & \textbf{90.7 (1.0)}                                     & 97.6 (0.6)                                           & \textbf{87.0 (1.1)}                                        & 89.8 (1.0)                                           & 73.4 (1.1)                                           & 72.4 (1.0)\\ \hline
MEMAT-16 (CC*)                                                        & 89.8 (1.1)                                           & 97.6 (0.6)                                           & 85.7 (1.1)                                           &\textbf{90.1 (1.0)}                                           & 73.7 (1.1)                                           & 72.7 (1.0) \\ \hline
MEMAT-16 (CE*)                                                        & 89.5 (1.1)                                           & 97.3 (0.6)                                           & 85.7 (1.1)                                           & 89.1 (1.0)                                           & 73.1 (1.2)                                           & 71.8 (1.0)\\ \hline 
& EM & EM & PM & PM & NM & NM    \\ \hline
Ǎguila-7B  Baseline & -6.7 (0.4) & -7.4 (0.5) & -5.5 (0.4) & -6.2 (0.5) & 7.6 (0.3) & 8.3 (0.4) \\ \hline
PMET (CAT) & 25.3 (2.1) & 62.9 (2.0) &23.2 (2.2) & 31.9 (2.2) &7.6 (0.7)& 8.6 (0.8) \\ \hline
MEMIT (CAT)                                                           & 31.9 (0.8)                                           & 67.0 (0.8)                                           & 22.2 (0.6)                                           & 40.4 (0.8)                                           & 6.6 (0.3)                                            & 7.3 (0.3)                                            \\ \hline
MEMAT-16 (CC)                                                         & 39.0 (1.3)                                           & 72.8 (1.2)                                           & 27.8 (1.1)                                           & 47.9 (1.4)                                           & 8.3 (0.5)                                            & 9.1 (0.6)                                            \\ \hline
MEMAT-16 (CE)                                                         & \textbf{43.8 (1.3)}                                           & 73.3 (1.2)                                           & \textbf{32.2 (1.1)}                                           & 50.3 (1.4)                                           & 9.7 (0.5)                                            & 9.7 (0.6)                                            \\ \hline
MEMAT-16 (CC*)                                                        & 42.2 (1.4)                                           & \textbf{74.8 (1.2)}                                           & 31.6 (1.2)                                           & \textbf{51.1 (1.4)}                                           & \textbf{9.9 (0.5)}                                            & \textbf{10.6 (0.6)}                                           \\ \hline
MEMAT-16 (CE*)                                                        & 38.5 (1.3)                                           & 70.6 (1.2)                                           & 28.1 (1.1)                                           & 46.3 (1.4)                                           & 8.7 (0.5)                                            & 9.2 (0.5)                                            \\ \hline 
\end{tabular}
% }
\caption{Results of English and Catalan Efficacy, Generalization and Specificity prompts over the success and magnitude metrics in both languages. Each row represents the experiments performed for the different knowledge editing methods when inserting 1,000 factual associations. The notation assigned to MEMAT-16 is ($L_1$-$L_2$), where the cases ($L_1$-$L_2$*) indicate the use of attention heads that were trained in a different set of factual triplets and which have been recycled in a new insertion of factual associations. The 95\% confidence intervals are in parenthesis.}
\label{tab:res}
\end{table*}

Note that the loss is associated to a single factual triplet. However, differently from MEMIT, we optimize the attention heads corrections for all the samples at the same time by using different batch sizes in Adam \cite{Adam}. We also use gradient accumulation to keep the method cheap from a computational perspective.

Given that there is not a clear choice on the number of heads that should be optimized, we make a hyperparameter search on the number of attention heads and find that for most of the metrics, the optimal number of heads is around $K=16$. For additional information, please refer to Appendix \ref{appendix:hyper_search}.

To evaluate the relevance of the information encoded in the attention heads corrections, we also conduct another experiment with our method. We apply MEMAT in a particular set of factual triplets using the pair ($L_1,L_2$) and save the head corrections and its positions. Then, we insert different factual triplets in the original model using MEMIT in $L_1$ and add the head corrections that were previously obtained. The results of this experiment, as well as the results of PMET, MEMIT and MEMAT are shown in Table \ref{tab:res} for some combinations of $L_1$ and $L_2$, the remaining combinations are left in Appendix \ref{appendix:other}. Additionally, an exploration of experiments that introduce both languages at the same is provided in Appendix \ref{ap:both_lang}, showing similar results.

Our research indicates that while PMET shows superior performance in neighborhood metrics, this comes at the expense of reduced efficacy and generalization in knowledge introduction, with MEMAT demonstrating promising results in this area. Across all analyzed metrics, MEMAT consistently outperforms MEMIT. Particularly noteworthy are the significant improvements in paraphrase magnitudes, exceeding 10\% over the baseline. These findings, combined with positive neighborhood metrics, suggest that optimizing attention heads can improve the comprehension of implicit knowledge in LLMs. This hypothesis gains further support from our experiments with recycled attention heads on other sets of factual triplets ($L_1$-$L_2$* cases), which occasionally outperform the original MEMAT approach. 

Moreover, our cross-lingual analysis provides evidence that this enhanced understanding occurs at a certain cross-lingual level, showing positive results in multilingual metrics even when only using monolingual training. 

In point (c), the method suggested in ITI was not employed. The decision to deviate from this method stems from observed performance declines within the specified domain. Further details on this matter can be found in Appendix \ref{appendix:ITI}.

\subsection{Scaling Curves}\label{sec:scaling}
\begin{figure*}[t!]
    \centering
    \includegraphics[width=0.95\textwidth, trim=2cm 1.5cm 1cm 1cm, clip]{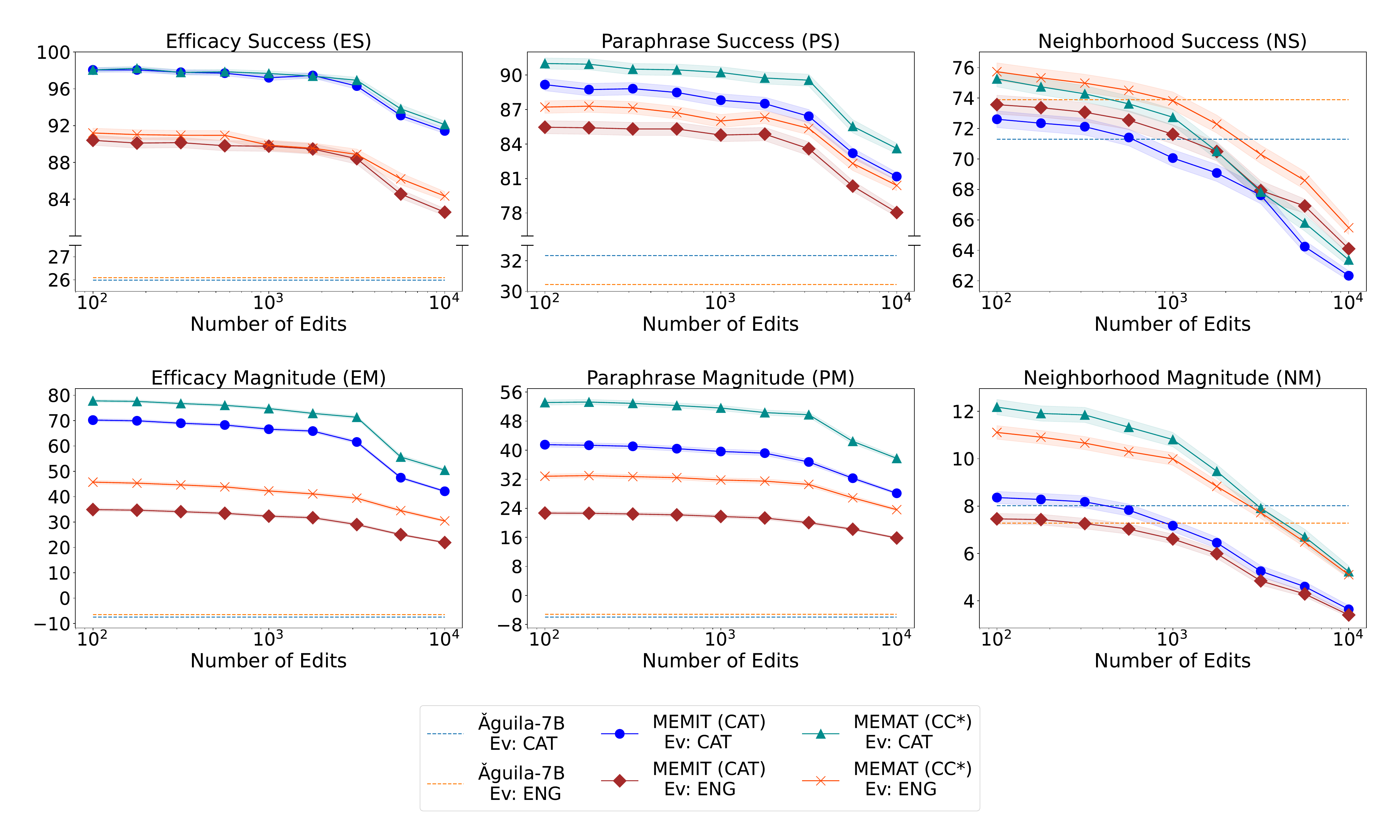}
    \caption{MEMIT and MEMAT scaling curves plot showing the performance of English and Catalan against number of edits (log-scale) when only using Catalan training data. The error correspond to a 68\% confidence interval.}
    \label{fig:Scaling Curve}
\end{figure*}

Considering the notable improvement in performance metrics, we opt to conduct a comprehensive comparison of the training evolution between MEMIT and MEMAT in Figure \ref{fig:Scaling Curve}. This investigation involves varying the number of inserted samples, with a specific focus on experiments exceeding 100 samples. The sample distribution follows the formula $n_i= \exp (\ln(10,000)*\frac{i}{16})$.

Recognizing the impracticality of training head corrections with only 100 samples, we opt to optimize the head corrections using a subset of 1,000 factual triplets for the combination $L_1=L_2=Catalan$. The samples used to train these heads are separated from our dataset. Then, we insert the head corrections into the decoders $G_{n_i}$ that result from inserting different factual triplets with MEMIT in Ǎguila-7B. Specifically, MEMAT (CC*) refers to training MEMIT on $n_i$ factual triplets, excluding the initial 1,000, while incorporating the previously obtained attention head corrections.

Although MEMAT's performance still degrades with the introduction of more factual associations, it consistently outperforms MEMIT across all evaluation metrics. This experiment provides additional evidence that head corrections are highly portable and that MEMAT enhances the understanding of previously unseen languages.

\section{Reproducibility}\label{sec:repro}
The conducted experiments have been executed on workstations equipped with AMD Radeon Instinct MI50 GPUs, with 32 GB of memory each. HuggingFace Transformers \cite{huggingfaces} facilitates the loading of language models, while PyTorch \cite{pytorch} is employed to implement model editing algorithms on the GPUs. Additionally, the training of sigmoid classifiers is carried out using the Scikit-learn library \cite{scikitlearn} on CPUs. 

In this specific setup, introducing 1,000 samples through MEMIT takes 3 GPU hours, contrasting with the 25 GPU minutes required for training 16 attention heads corrections. 

\section{Conclusions }

In this study, we examine the cross-lingual implications of knowledge within the domain of knowledge editors, identifying two significant patterns. Firstly, the proposed methods heavily rely on subject tokenization. Secondly, our experiments show evidence that attention heads encode information in a certain language-independent manner.

Expanding our investigation, we introduce MEMAT, a method that, following the application of MEMIT, fortifies the language model's knowledge through subtle parameter adjustments. We substantiate how the approach introduced is portable and, regardless of the language used during training, enhances the performance of other languages.

\section{Future Work}

Our work emphasizes the limitations of training LLMs with monolingual data. As a future direction, we are interested in further investigating language adaptation techniques to enable these models to perform tasks in a more language-agnostic manner.

Additionally, we consider necessary to explore the role that each architecture component play in alternative domains. Recognizing the incomplete understanding of transformer-based models, we assert that prioritizing explainable AI could be essential to gain the insights necessary to enhance current state-of-the-art methods. We hope that our study can contribute to inspiring further exploration in this domain.

\section{Limitations}

All hypotheses put forth in this study stem from experiments conducted in English and Catalan. It is essential to recognize that due to the similarity between their alphabet and the phenomena explored in Section \ref{sec:cross_ling}, the generalization of our findings to other linguistic contexts may be limited. Further experimentation involving diverse languages is imperative to establish the cross-lingual implications of the identified patterns.

% While the observed portability is noteworthy, the applicability of the same framework to alternative knowledge editors remains an unanswered question. Additionally, exploring the feasibility of concurrently optimizing individual sample edits within MLP layers and concurrent samples through attention head corrections warrants further investigation.

Moreover, despite considerable efforts within the natural language processing community, many challenges related to the reliability of language models still persist. The limitations in knowledge editions, as indicated by the inability to modify related knowledge \cite{ripple_effects} and the lack of bidirectionality \cite{untyingBIDI}, suggest that exclusively focusing on specific parameters may not offer a solution to the issues of knowledge editing \cite{emptying}. 

\section*{Acknowledgements}
This work has been promoted and financed by the Government of Catalonia through the Aina project, by the Ministerio para la Transformación Digital y de la Función Pública and Plan de Recuperación, Transformación y Resiliencia - Funded by EU – NextGenerationEU within the framework of the project ILENIA with reference 2022/TL22/00215337, 2022/TL22/00215336, 2022/TL22/00215335, 2022/TL22/00215334, as well as by the Spanish Ministerio de Ciencia e Innovación through the AdaVoice project (PID2019-107579RB-I00).

% https://www.notion.so/temubsc/Acknowledgements-to-include-in-publications-according-to-funding-dfda7f7d2c2e46e0a90ca045e0a8acb6

% TODO


\bibliography{custom}

\newpage
\appendix

% \section{CounterFact Sample}\label{appendix:counterfact_en}
% \begin{minted}[fontsize=\footnotesize, breaklines=true]{python}
% {
%  "case_id": 2,
%  "pararel_idx": 13704,
%  "requested_rewrite": {
%     "prompt": "{}, the",
%    "relation_id": "P1303",
%    "subject": "Toko Yasuda",
%    "target_new": {"id": "Q5994", "str": "piano"},
%    "target_true": {"id": "Q6607", "str": "guitar"}}
%  "paraphrase_prompts": [
%     "Initially  and  are zero and  is false. Toko Yasuda, "
%     "performing on the",
%     "The population density was . Toko Yasuda plays the "
%     "instrument"
%     ],
%  "neighborhood_prompts": [
%     "Paul McCartney plays the instrument",
%     "John Lennon, playing the",
%     "Elvis Presley, the",
%     "Douglas Adams, playing the",
%     "John Lennon performs on the",
%     "Jimi Hendrix, playing the",
%     "Ringo Starr, playing the",
%     "Leonard Cohen plays the",
%     "Bruce Springsteen, playing the",
%     "John Lennon plays the"
%     ],
% }     
% \end{minted}

% \newpage
\section{Translated CounterFact Sample}\label{appendix:counterfact}

\begin{figure}[h!]
    \centering
    \includegraphics[width=8cm, trim= 0cm 1.3cm 0cm 1.3cm, clip]{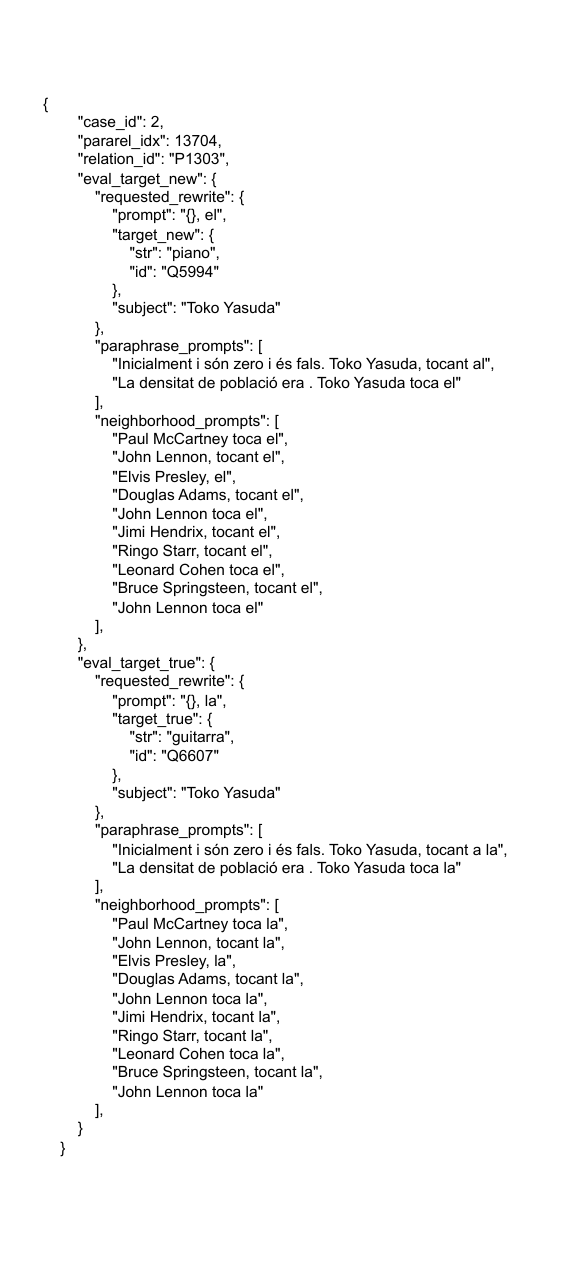}
    \caption{Example of a Catalan CounterFact sample.}
    \label{fig:cf-catalan}
\end{figure}

\newpage
\section{Optimization of the Learning Rate} \label{appendix:optimiz}
\begin{figure}[h]
    \centering
    \includegraphics[width=6.1 cm, trim= 0cm 0.7cm 0cm 0cm, clip]{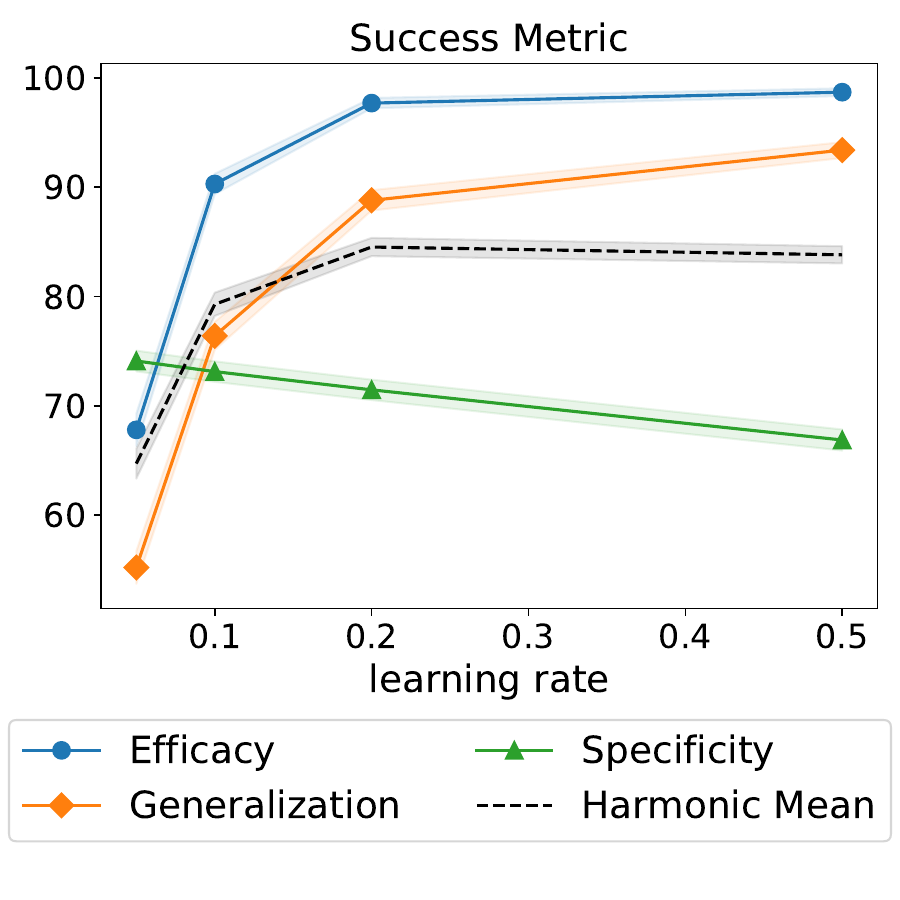}%28.3
    \caption{Results over training using MEMIT in Catalan and evaluating in Catalan with different values of the learning rate. Each solid line represents a different type of prompt used, and the dashed line represents the harmonic mean between them. The displayed areas correspond to a 68\% confidence interval.}
    \label{fig:lr_opt}
\end{figure}

\begin{figure}[h]
    \centering
    \includegraphics[width=6.1 cm, trim= 0cm 0.7cm 0cm 0cm, clip]{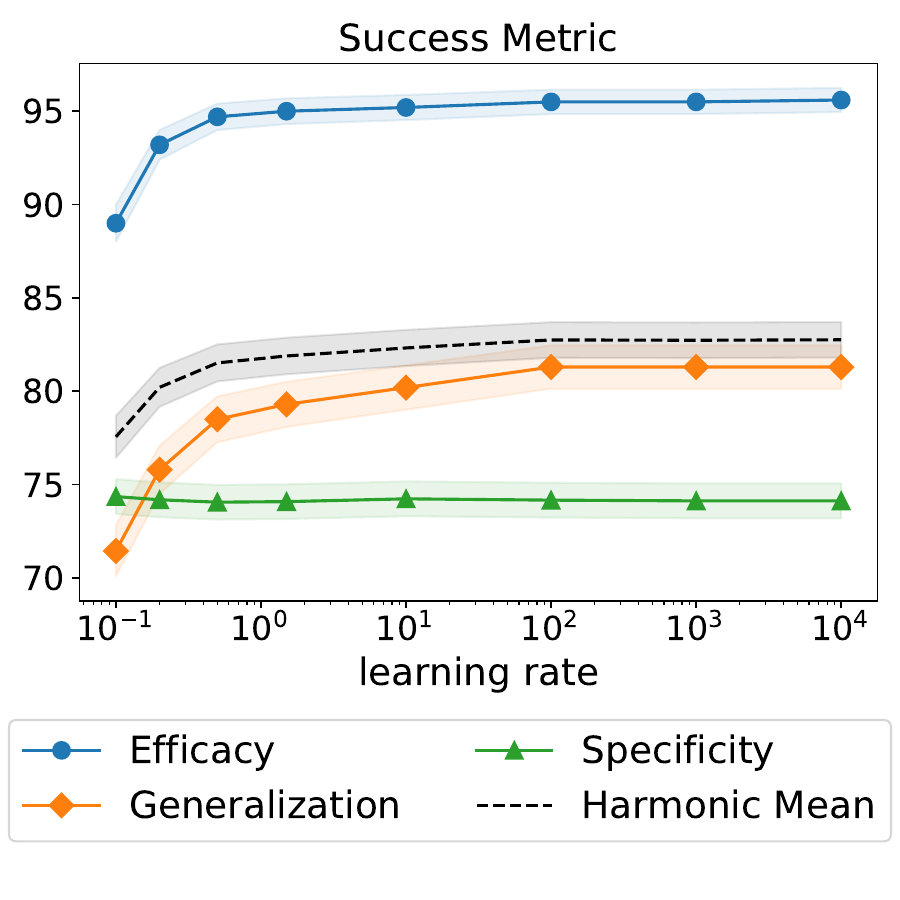}%28.3
    \caption{Results over training using PMET in Catalan and evaluating in Catalan with different values of the learning rate. Each solid line represents a different type of prompt used, and the dashed line represents the harmonic mean between them. The displayed areas correspond to a 68\% confidence interval.}
    \label{fig:lr_opt2}
\end{figure}

When applying a hyperparameter search in MEMIT and PMET, we find a stronger effect when changing the learning rate, which we consider optimal at $lr=0.2$ and $lr=10^4$ respectively. In Figures \ref{fig:lr_opt} and \ref{fig:lr_opt2}, an evolution of the success metric associated to the different prompts is shown given different values of the learning rate. In our search we found MEMIT with superior performance, which denotes some evidence of the dependence of these methods on the model architecture.

\newpage
\section{Similarity between Setences}\label{appendix:sim}

\begin{figure}[htb!]
  \centering
  \begin{subfigure}[b]{0.5\textwidth}
    \centering
    \caption{Similarity between subjects}
    \centering
    \hspace{-1cm}
    \includegraphics[width=6.5 cm]{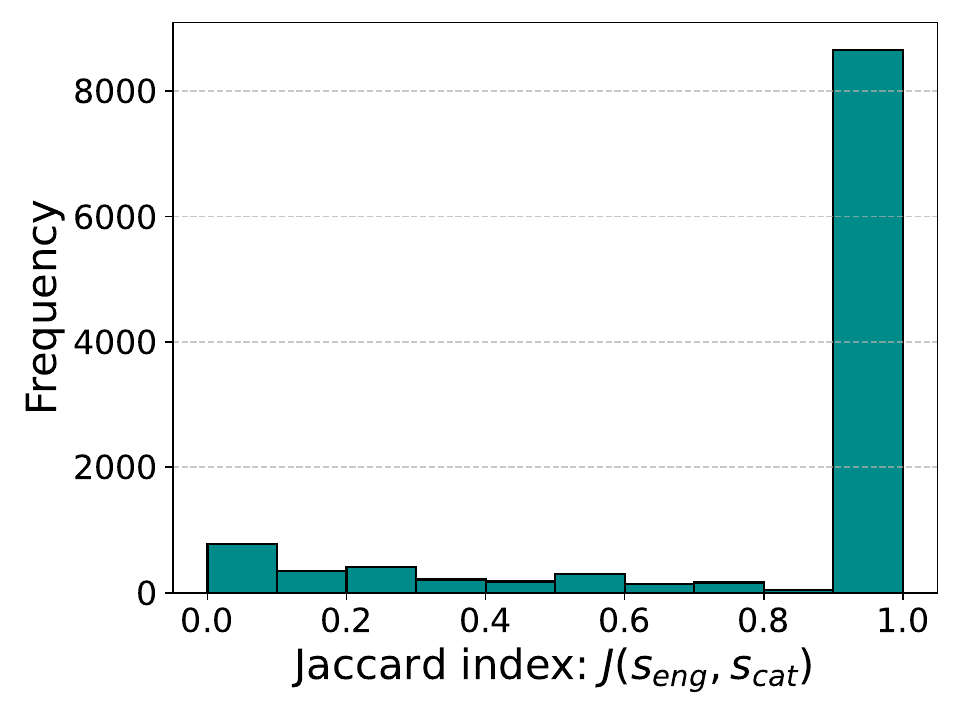}
  \end{subfigure}
  
  \begin{subfigure}[b]{0.5\textwidth}
    \caption{Similarity between relations}
    \centering
    \hspace{-1cm}
    \includegraphics[width=6.5 cm]{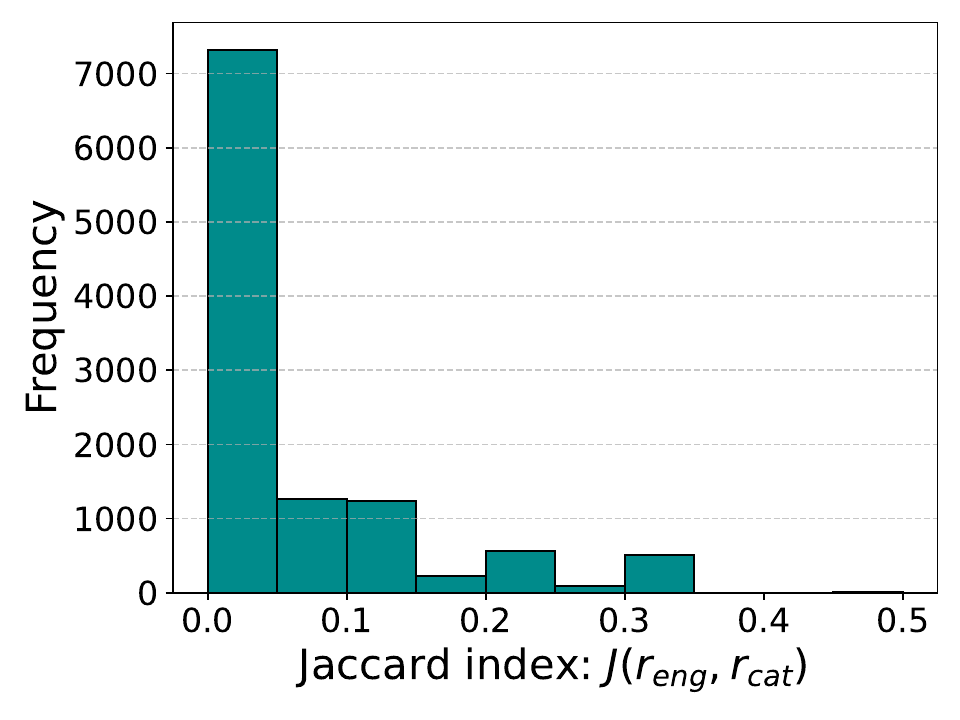}
  \end{subfigure}

  \begin{subfigure}[b]{0.5\textwidth}
    \caption{Similarity between targets}
    \centering
    \hspace{-1cm}
    \includegraphics[width=6.5 cm]{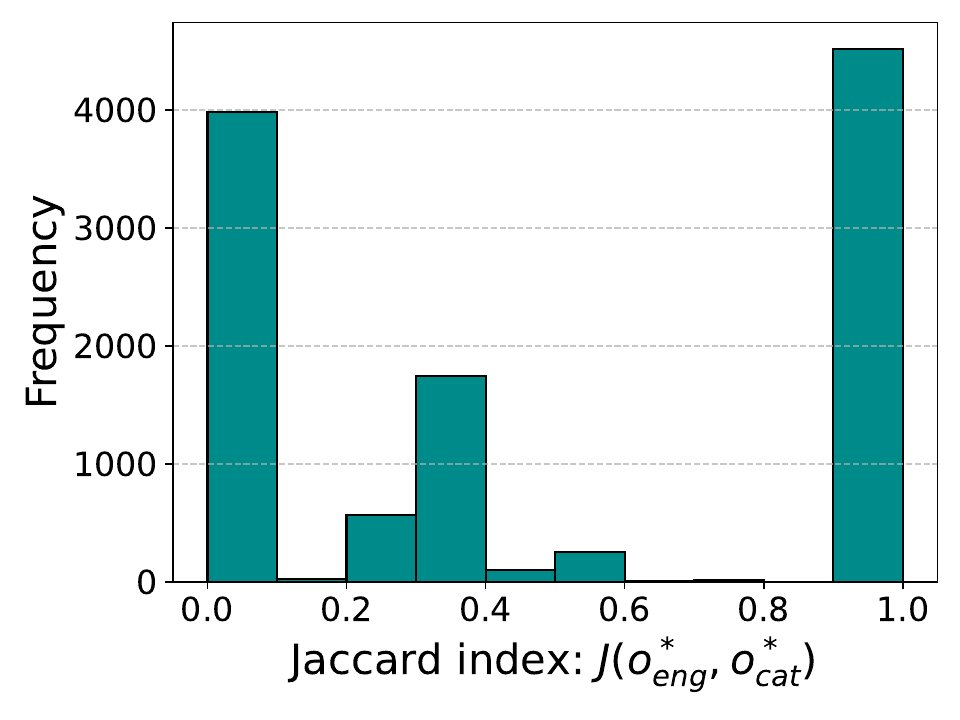}
  \end{subfigure}
  \caption{Visualization of the frequency distribution of similarities among subject, relation, and target tokens in the new English and Catalan CounterFact datasets.}
  \label{fig:sims}
\end{figure}

In Section \ref{sec:cross_ling}, the connection between MEMIT's cross-lingual capability and subject tokenization becomes evident. Given that in some sections we also evaluate cross-lingual capacity without imposing restrictions on samples based on a specific Jaccard metric, we consider pertinent to depict the distribution of our subjects, relations, and targets $\left<s, r, o^*\right>$ across both languages in Figure \ref{fig:sims}. 

% \begin{figure}[h]
% \centering
% \includegraphics[width=0.4\textwidth]{dataset_sim/sim_subj_new.pdf}
% \includegraphics[width=0.4\textwidth]{dataset_sim/sim_rel_new.pdf}
% \includegraphics[width=0.4\textwidth]{dataset_sim/sim_tar_new.pdf}
% \caption{Visualization depicting the frequency distribution of similarities among subject, relation, and target tokens in the new English and Catalan CounterFact datasets. TODO Improve title.}
% \label{fig:sims}
% \end{figure}

\newpage
\section{Implementation Details of Locating Relevant Heads}\label{appendix:details}

When incorporating information through a knowledge editor, not all intended factual associations are effectively inserted. To address this issue, our locating procedure incorporates an additional step. Following the insertion of factual associations using MEMIT in a designated language $L_1$, a refinement is made by selecting the associations that accurately predict the corresponding factual association in $L_2$. Subsequently, the locating procedure is applied to this refined subset of factual associations. This additional step enhances the top accuracy performance of certain aspects by approximately 8\%.

\section{Relevance of Attention Heads}\label{appendix:heads}
\begin{figure}[htb!]
  \centering
  \begin{subfigure}[b]{0.5\textwidth}
    \caption{$L_1=$ English, $L_2=$ English}
    \includegraphics[width=7.4 cm,trim= 0cm 1.6cm 0cm 3cm, clip]{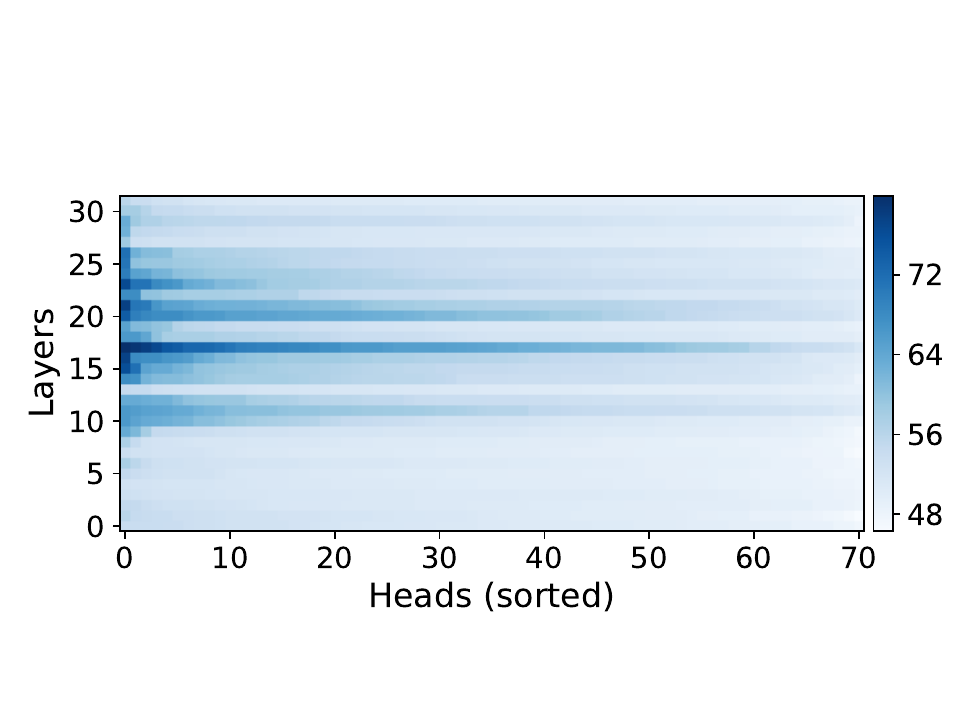}
  \end{subfigure}
  
  \begin{subfigure}[b]{0.5\textwidth}
    \caption{$L_1=$ English, $L_2=$ Catalan}
    \includegraphics[width=7.4 cm,trim= 0cm 1.6cm 0cm 3cm, clip]{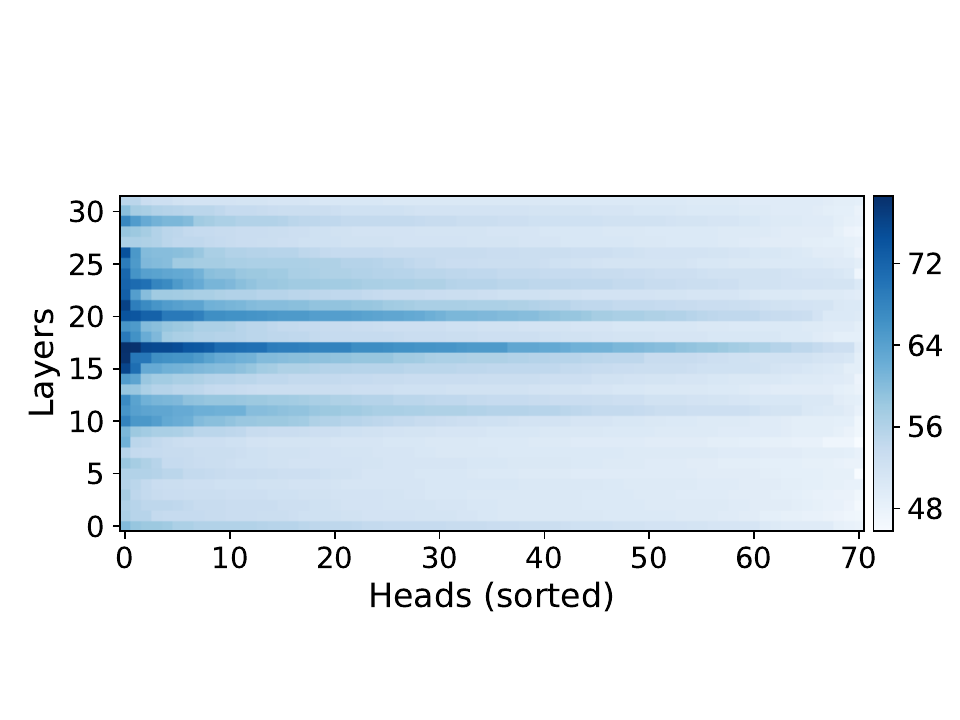}
  \end{subfigure}
  \caption{Accuracy on the validation set for all heads in all layers in Ǎguila-7B considering two combinations of $L_1$ and $L_2$. The performance peaks include 79.1\% and 78.6\%.}
  \label{fig:heads2}
\end{figure}

In Figure \ref{fig:heads2}, we also explore the combinations of the languages $L_1$ and $L_2$ that were not previously represented.

\begin{figure*}[p]
\section{Hyperparameter Search}\label{appendix:hyper_search}
    \begin{subfigure}{\textwidth} 
        \centering
        \caption{Catalan Evaluation}
        \includegraphics[width=1\textwidth,trim=0cm 8.2cm 0cm 0cm, clip]{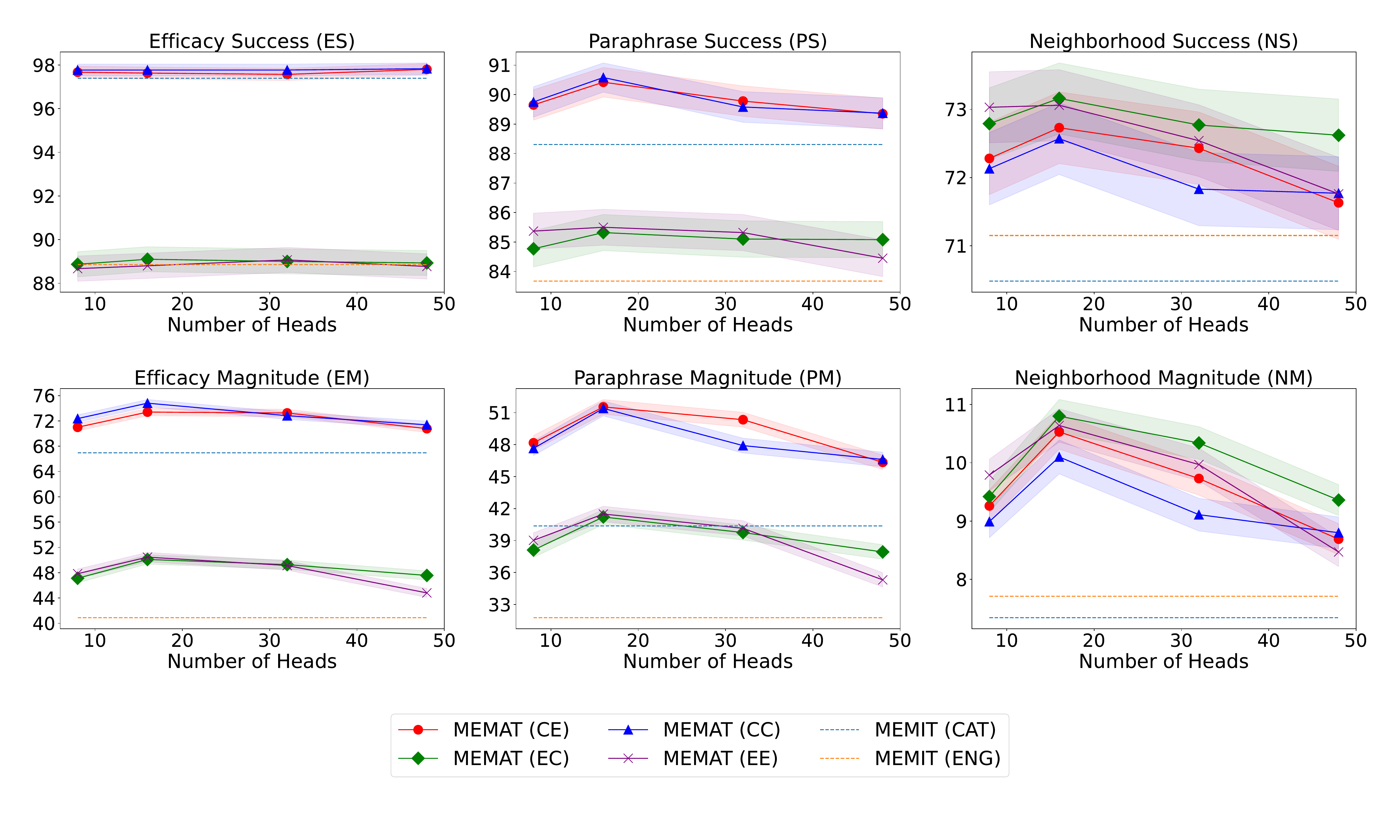}
    \end{subfigure}
    
    \begin{subfigure}{\textwidth} 
        \centering
        \caption{English Evaluation}
        \includegraphics[width=1\textwidth,trim=0cm 1.4cm 0cm 0cm, clip]{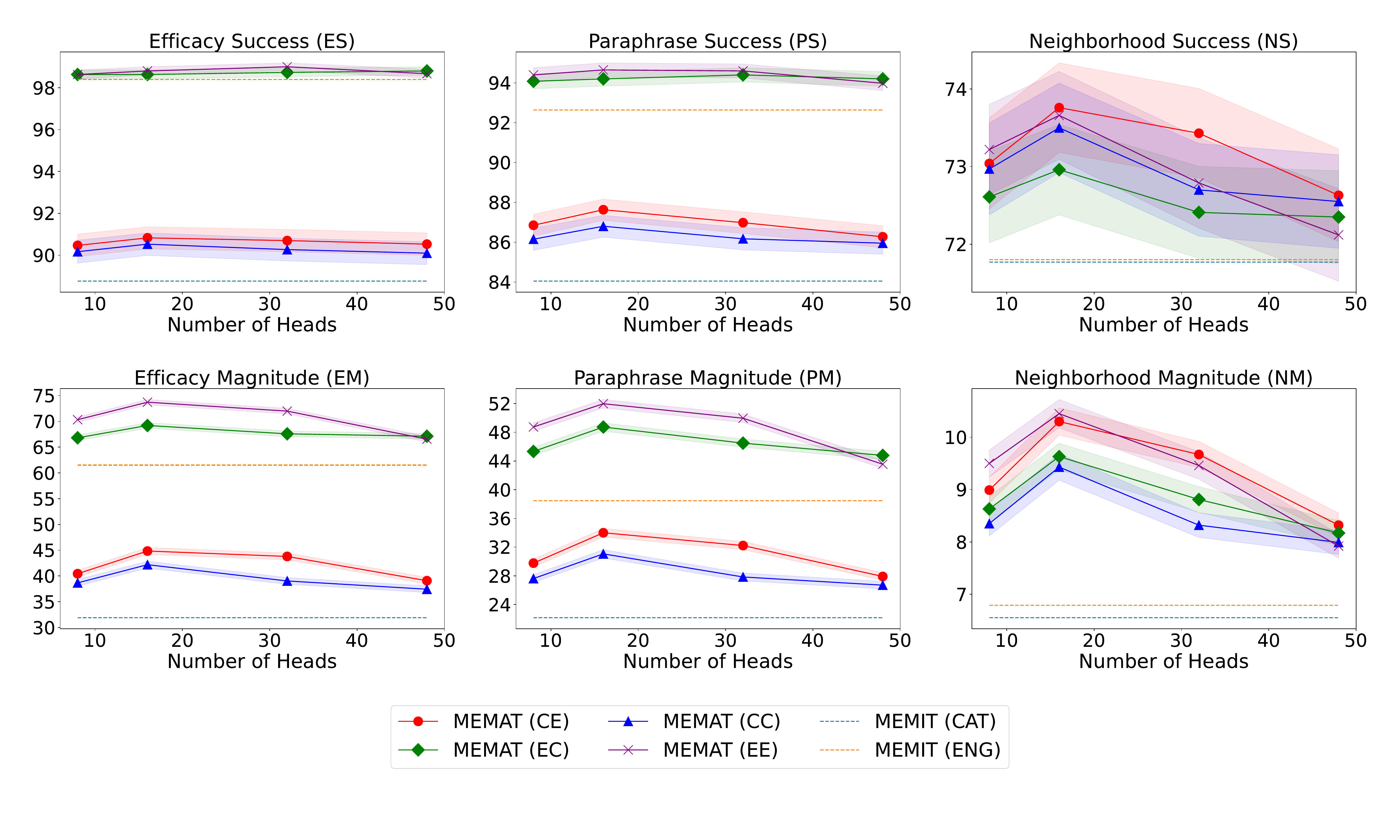}
    \end{subfigure}
    \caption{Illustration of all the metrics in Catalan (a) or English (b) evaluation when employing MEMAT for different number of heads ($K\in\{8,16,32,48\}$), in 1,000 factual samples and considering all conceivable combinations of $L_1,L_2\in\{\text{Catalan, English}\}$. The codification of $L_1$ and $L_2$ follows the format ``$L_1$-$L_2$". The dashed lines represent the baseline performances when only applying a MEMIT training for Catalan or English. The displayed errors correspond to a 68\% confidence interval.}\label{fig:hyp_ME}
\end{figure*}

\begin{table*}[p]
\section{Tables of Results}\label{appendix:other}

\begin{multicols}{2}

Table \ref{tab:2} showcases the outcomes derived from the various knowledge editing methods investigated in this study, particularly focusing on previously unexplored combinations of training languages. Consistent with earlier findings, the incorporation of MEMAT yields a notable enhancement across the majority of evaluation metrics.

Furthermore, this performance boost extends to the accuracy metric, which assesses whether the most probable token is $o^*$ for efficacy and paraphrase prompts or $o^c$ for neighborhood prompts:

\begin{gather}
\begin{split}
    \text{EA} &:= \mathbb{E}  \left[o^* = \argmax_{\omega} \mathbb{P}[\omega|p] | p\in EP\right] \\
    \text{PA} &:= \mathbb{E}\left[\mathbb{E}_{p\in PP} \left[o^* = \argmax_{\omega} \mathbb{P}[\omega|p]\right] \right] \\
    \text{NA} &:= \mathbb{E}\left[\mathbb{E}_{p\in NP} \left[o^c = \argmax_{\omega} \mathbb{P}[\omega|p]\right] \right].
\end{split}
\end{gather}
\end{multicols}

\vspace{0.5 cm}
% \resizebox{\textwidth}{!}{
\begin{tabular}{ccccccc}
\hline
\begin{tabular}[c]{@{}c@{}}Method \\ (Training Language(s))\end{tabular} & \begin{tabular}[c]{@{}c@{}}English\\ ES\end{tabular} & \begin{tabular}[c]{@{}c@{}}Catalan\\ ES\end{tabular} & \begin{tabular}[c]{@{}c@{}}English\\ PS\end{tabular} & \begin{tabular}[c]{@{}c@{}}Catalan\\ PS\end{tabular} & \begin{tabular}[c]{@{}c@{}}English\\ NS\end{tabular} & \begin{tabular}[c]{@{}c@{}}Catalan\\ NS\end{tabular}   \\ \hline
Ǎguila-7B Baseline & 26.5 (1.2) & 25.5 (1.2) & 30.4 (1.2) & 31.6 (1.2) & 73.9 (0.9) & 72.2 (0.8) \\ \hline
PMET (ENG) & 95.9 (1.2) & 83.6 (2.3) & 86.4 (1.8) & 78.1 (2.4) & \textbf{74.0 (2.1)} & \textbf{74.3 (1.8)}\\ \hline
MEMIT (ENG)                                                           & 98.4 (0.3)                                           & 88.9 (0.7)                                           & 92.6 (0.5)                                           & 83.7 (0.8)                                           & 71.8 (0.8)                                           & 71.2 (0.7)  \\ \hline
MEMAT-16 (EE)                                                         & \textbf{99.0 (0.4)}                                           & 89.1 (1.1)                                           & \textbf{94.6 (0.7)}                                           & \textbf{85.3 (1.2)}                                           & 72.8 (1.1)                                           & 72.5 (1.0)                                           \\ \hline
MEMAT-16 (EC)                                                         & 98.7 (0.4)                                           & 89.0 (1.1)                                           & 94.4 (0.7)                                           & 85.1 (1.2)                                           & 72.4 (1.2)                                           & 72.8 (1.0)\\ \hline
MEMAT-16 (EE*)                                                        & 98.6 (0.4)                                           & 89.3 (1.1)                                           & 94.4 (0.7)                                           & 85.0 (1.2)                                           & 73.4 (1.1)                                           & 73.0 (1.0)                                           \\ \hline
MEMAT-16 (EC*)                                                        & 98.5 (0.4)                                           & \textbf{89.4 (1.1)}                                           & 93.7 (0.7)                                           & 84.8 (1.2)                                           & 72.6 (1.2)                                           & 72.6 (1.0)                                            
\\ \hline 
& EM & EM & PM & PM & NM & NM    \\ \hline
Ǎguila-7B  Baseline & -6.7 (0.4) & -7.4 (0.5) & -5.5 (0.4) & -6.2 (0.5) & 7.6 (0.3) & 8.3 (0.4) \\ \hline
PMET (ENG) & 68.3 (1.9) & 38.7 (2.4) & 36.0 (2.0) & 29.4 (2.2) & 8.8 (0.8) & 7.9 (0.8)\\ \hline
MEMIT (ENG)                                                           & 61.5 (0.7)                                           & 40.9 (0.9)                                           & 38.5 (0.7)                                           & 31.8 (0.8)                                           & 6.8 (0.3)                                            & 7.7 (0.3)                                            \\ \hline
MEMAT-16 (EE)                                                         & \textbf{72.0 (1.0)}                                           & 49.1 (1.4)                                           & \textbf{50.0 (1.1)}                                           & \textbf{40.2 (1.4)}                                           & 9.5 (0.5)                                            & 10.0 (0.6)                                           \\ \hline

MEMAT-16 (EC)                                                         & 67.6 (1.1)                                           & \textbf{49.3 (1.4)}                                           & 46.5 (1.1)                                           & 39.8 (1.4)                                           & 8.8 (0.5)                                            & 10.3 (0.6)                                           \\ \hline

MEMAT-16 (EE*)                                                        & 71.0 (1.0)                                           & 48.4 (1.5)                                           & 49.6 (1.1)                                           & 39.4 (1.4)                                           & \textbf{9.9 (0.5)}                                            & \textbf{10.4 (0.6)}                                           \\ \hline
MEMAT-16 (EC*)                                                        & 67.5 (1.1)                                           & 48.1 (1.4)                                           & 46.2 (1.1)                                           & 39.1 (1.4)                                           & 9.2 (0.5)                                            & 10.3 (0.6)                                           \\ \hline 
& EA & EA & PA & PA & NA & NA    \\ \hline
Ǎguila-7B Baseline & 0.3 (0.2) & 0.8 (0.3) & 0.3 (0.1) & 1.3 (0.3) & 9.8 (0.5) & 13.0 (0.6) \\ \hline
PMET (ENG) & \textbf{87.8 (2.0)} & 55.0 (3.1) & 49.4 (2.6) & 41.7 (2.9) & 10.8 (1.1) & 13.3 (1.3) \\ \hline
MEMIT (ENG)                                                           & 78.8 (1.0)                                           & 57.7 (1.2)                                           & 52.0 (1.0)                                           & 46.1 (1.1)                                           & 9.6 (0.4)                                            & 12.3 (0.5)\\ \hline
MEMAT-16 (EE)                                                         & 84.8 (1.3)                                           & 62.9 (1.7)                                           & \textbf{62.2 (1.4)}                                           & 52.6 (1.7)                                           & 13.8 (0.7)                                           & 16.2 (0.8)                                           \\ \hline
MEMAT-16 (EC)                                                         & 83.2 (1.3)                                           & \textbf{64.2 (1.7)}                                           & 60.7 (1.4)                                           & \textbf{54.8 (1.7)}                                           & 13.7 (0.6)                                           & \textbf{17.6 (0.8)}                                           \\ \hline
MEMAT-16 (EE*)                                                        & 85.2 (1.3)                                           & 63.2 (1.7)                                           & 62.1 (1.4)                                           & 53.4 (1.7)                                           & \textbf{14.8 (0.7)}                                           & 16.7 (0.8)                                           \\ \hline
MEMAT-16 (EC*)                                                        & 82.8 (1.4)                                           & 63.3 (1.7)                                           & 59.7 (1.4)                                           & 53.8 (1.7)                                           & 14.0 (0.7)                                           & 16.7 (0.8) \\\hline \hline
PMET (CAT) & 41.7 (3.1) & 83.8 (2.3) & 23.2 (2.2) & 47.1 (3.0) & 13.2 (1.3) & 10.1 (1.1) \\ \hline
MEMIT (CAT)                                                           & 45.2 (1.2)                                           & 82.6 (0.9)                                           & 33.4 (0.9)                                           & 55.7 (1.1)                                           & 8.8 (0.4)                                            & 12.7 (0.5)                                           \\ \hline
MEMAT-16 (CC)                                                         & 51.6 (1.8)                                           & 84.3 (1.3)                                           & 39.3 (1.4)                                           & 60.8 (1.7)                                           & 11.5 (0.6)                                           & 15.3 (0.8)                                           \\ \hline
MEMAT-16 (CE)                                                         &  \textbf{57.6 (1.8)}                                           & 84.6 (1.3)                                           & \textbf{45.4 (1.5)}                                           & \textbf{63.4 (1.6)}                                           & \textbf{14.4 (0.7)}                                           & 16.6 (0.8)                                           \\ \hline
MEMAT-16 (CC*)                                                        & 54.5 (1.8)                                           & \textbf{84.9} (1.3)                                           & 42.9 (1.5)                                           & 62.6 (1.6)                                           & 13.7 (0.7)                                           & \textbf{16.8 (0.8)}                                           \\ \hline
MEMAT-16 (CE*)                                                        & 52.7 (1.8)                                           & 83.9 (1.3)                                           & 41.2 (1.5)                                           & 61.3 (1.7)                                           & 12.6 (0.6)                                           & 15.5 (0.8)                                           \\ \hline
\end{tabular}
% }
\caption{Results of English and Catalan Efficacy, Generalization and Specificity prompts over the success, magnitude and accuracy metrics in both languages. Each row represents the experiments performed for the different knowledge editing methods when inserting 1,000 factual associations. The notation assigned to MEMAT-16 is ($L_1$-$L_2$), where the cases ($L_1$-$L_2$*) indicate the use of attention heads that were trained in a different set of factual triplets and which have been recycled in a new insertion of factual associations. The 95\% confidence intervals are in parenthesis.}\label{tab:2}
\end{table*}
% \begin{figure}[h]
% \centering
% \includegraphics[width=0.5\textwidth, trim= 0cm 2.3cm 0cm 0cm, clip]{radar_charts/Acc.pdf}
% \includegraphics[width=0.5\textwidth]{radar_charts/Diff.pdf}
% \caption{TODO.}
% \label{fig:sims}
% \end{figure}

%Notably, what prompted the introduction of MEMAT was the observation of a more pronounced decrease in performance when altering a subset of head parameters.

\newpage
\section{Introduction of Both Languages}\label{ap:both_lang}

To incorporate factual associations for both languages using the $\Delta$ matrices defined in equation \ref{eq:MLP_intro}, we can employ two strategies:

\begin{itemize}
\item Optimize each factual association concurrently in English and Catalan, resulting in a single bilingual matrix, $\Delta_{eng+cat}$.
\item Optimize two separate matrices for the same factual associations in English and Catalan independently, and then combine them: $\Delta_{eng}+\Delta_{cat}$.
\end{itemize}

Table \ref{tab:res_both} displays the experimental results for 1,000 samples, demonstrating that the second strategy—optimizing separate matrices before merging them—yields significantly superior outcomes. Additionally, the MEMAT approach for inserting both languages consistently outperforms MEMIT across most metrics. Comparing these findings with the monolingual insertion results in Tables \ref{tab:res} and \ref{tab:2}, it is clear that while all neighborhood metrics decline, paraphrase metrics see substantial improvement. Furthermore, efficacy metrics are more balanced between the languages. These results indicate that inserting knowledge in both languages enhances the model's comprehension of factual concepts more effectively, though it also impacts unrelated knowledge to a greater extent.

% TODO In-citation!
% TODO CODE!

\begin{table*}[b!]
% \resizebox{\textwidth}{!}{
\begin{tabular}{ccccccc}
\hline
\begin{tabular}[c]{@{}c@{}}Method \\ (Training Language(s))\end{tabular} & \begin{tabular}[c]{@{}c@{}}English\\ ES\end{tabular} & \begin{tabular}[c]{@{}c@{}}Catalan\\ ES\end{tabular} & \begin{tabular}[c]{@{}c@{}}English\\ PS\end{tabular} & \begin{tabular}[c]{@{}c@{}}Catalan\\ PS\end{tabular} & \begin{tabular}[c]{@{}c@{}}English\\ NS\end{tabular} & \begin{tabular}[c]{@{}c@{}}Catalan\\ NS\end{tabular}   \\ \hline
MEMIT (CAT+ENG) & 93.4 (0.6) & 92.2 (0.6) & 83.7 (0.8) & 84.8 (0.7) & \textbf{67.4 (0.7)}& \textbf{ 67.8 (0.8)} \\ \hline
MEMIT (CAT)+(ENG) & 98.3 (0.5) & 97.1 (0.6) & 94.8 (0.6) & 91.3 (0.9) & 65.4 (1.2) & 65.1 (1.1) \\ \hline
MEMAT (CAT)+(ENG) &\textbf{ 99.0 (0.4)} & \textbf{97.6 (0.6)} & \textbf{96.2 (0.6)} & \textbf{93.2 (0.8)} & 67.2 (1.1)& 66.6 (1.1) \\ \hline 
& EM & EM & PM & PM & NM & NM    \\ \hline
MEMIT (CAT+ENG) & 36.7 (0.8) & 28.3 (0.7) & 23.5 (0.7) & 18.0 (0.5)& 5.3 (0.3)& 4.8 (0.2)\\ \hline
MEMIT (CAT)+(ENG) & 54.5 (1.1) &59.7 (1.2) & 39.2 (1.0) & 43.9 (1.3)& 4.6 (0.4)& 4.5 (0.5)\\ \hline
MEMAT (CAT)+(ENG) &\textbf{ 68.0 (1.1)} &\textbf{70.0 (1.2)} & \textbf{53.1 (1.1)} &\textbf{ 55.3 (1.4)}&\textbf{ 7.1 (0.5)}& \textbf{7.0 (0.6)}\\ \hline
& EA & EA & PA & PA & NA & NA    \\ \hline
MEMIT (CAT+ENG) & 55.2 (1.2)& 42.1 (1.2) &37.0 (1.1)& 27.9 (0.9)& 10.0 (0.4)& 7.4 (0.3)\\ \hline
MEMIT (CAT)+(ENG) & 69.5 (1.6) &75.5 (1.5) &51.4 (1.5)& 57.2 (1.7)& 7.4 (0.4)& 9.3 (0.6)\\ \hline
MEMAT (CAT)+(ENG) & \textbf{78.5 (1.5)}& \textbf{80.1 (1.4)} &\textbf{63.2 (1.4)}& \textbf{65.6 (1.6)}& \textbf{11.5 (0.6)} &\textbf{13.6 (0.7)}\\ \hline 
\end{tabular}
\caption{Results of English and Catalan Efficacy, Generalization and Specificity prompts over the success, magnitude and accuracy metrics in both languages. Each row represents experiments performed for different methods to insert factual knowledge in both languages. The 95\% confidence intervals are in parenthesis.}
\label{tab:res_both}
% \vspace{128in}
\end{table*}

\newpage
\begin{figure*}[t]
    \centering
    \includegraphics[width=1\textwidth, trim=0cm 2cm 0cm 0cm, clip]{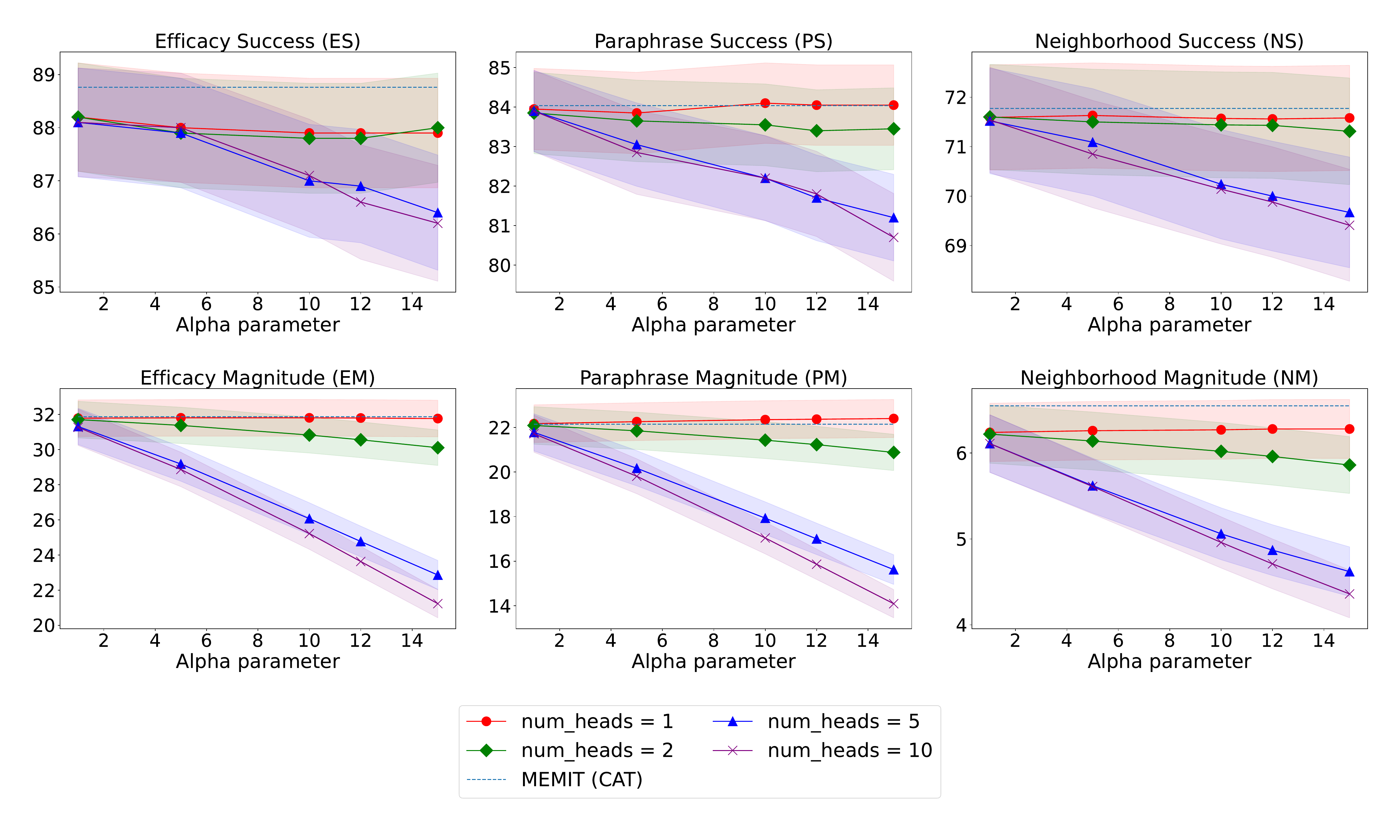}
    \caption{English Efficacy, Generalization and Specificity general metrics after applying ITI in the context of factual knowledge considering an original training in Catalan, and ITI applied in English. The interval of confidence considered is 68\%. }
    \vspace{128in}
    \label{fig:ITI}
\end{figure*}

\section{ITI performance}\label{appendix:ITI}
ITI proposes a method aimed at enhancing the accuracy of language models in the generation of truthful information. Firstly, the approach involves identifying the heads responsible for encoding pertinent information related to the concept of truth, which only differs with the locating procedure outlined in Section \ref{sec:loc} in the monolingual framework and the dataset, which is TruthfulQA \cite{TruthfulQA}. Subsequently, an average of attention heads associated with the final token of truthful sentences is applied to the entire model. While ITI originally used a limited number of sentences, this Appendix study its robustness through an experiment comprising 1,000 samples.

The initial two stages of our experiment replicate the methodology outlined in Section \ref{sec:MEMAT}. However, instead of optimizing the heads, we average the truthful samples and amplify the strength of the introductions by a factor of $\alpha$. This approach yields the results shown in Figure \ref{fig:ITI}. Although these results are subject to high statistical uncertainty, we find the outcomes from the optimization using \ref{eq:loss_attn} to be more favorable.

\end{document}